\documentclass{article}
\PassOptionsToPackage{numbers, compress}{natbib}
\usepackage[final]{neurips_2023}
\usepackage{microtype}
\usepackage{graphicx}
\usepackage{booktabs} 
\usepackage{hyperref}

\usepackage{tikz}
\usepackage[most]{tcolorbox}
\usetikzlibrary{fit,calc}
\newcommand{\boxit}[2]{
    \tikz[remember picture,overlay] \node (A) {};\ignorespaces
    \tikz[remember picture,overlay]{\node[yshift=3pt,fill=#1,opacity=.25,fit={($(A)+(0,0.15\baselineskip)$)($(A)+(.9\linewidth,-{#2}\baselineskip - 0.25\baselineskip)$)}] {};}\ignorespaces
}

\usepackage{adjustbox}
\usepackage{mathtools}
\usepackage[utf8]{inputenc} 
\usepackage[T1]{fontenc}    
\usepackage{url}            
\usepackage{amsfonts}       
\usepackage{nicefrac}       
\usepackage{xcolor}         
\usepackage{colortbl}
\usepackage{subcaption}
\usepackage[ruled,vlined]{algorithm2e}
\usepackage{algpseudocode}
\usepackage{enumitem}
\usepackage{multirow}
\usepackage{capt-of}
\usepackage{wrapfig}
\usepackage{makecell}
\usepackage{bm}
\usepackage{arydshln}
\usepackage{placeins}
\usepackage{amssymb}

\hypersetup{
    colorlinks = true,
    citecolor = blue,
    linkcolor = red
}
\usepackage{comment}
\usepackage{eqparbox}

\usepackage{adjustbox}
\usepackage{amsmath}
\DeclareMathOperator*{\argmax}{arg\,max}

\definecolor{gg}{gray}{0.92}
\newcolumntype{a}{>{\columncolor{gg}}c}
\definecolor{figred}{RGB}{255, 181, 164}
\definecolor{figblue}{RGB}{156,192,231}
\definecolor{figgreen}{RGB}{253, 229, 180}

\definecolor{fig2green}{RGB}{29, 120, 116}
\definecolor{fig2blue}{RGB}{161, 85, 185}
\definecolor{fig2yellow}{RGB}{246, 213, 92}
\definecolor{fig2red}{RGB}{238, 46, 49}

\usepackage[capitalize,noabbrev]{cleveref}
\usepackage[textsize=tiny]{todonotes}
\usepackage[hang,flushmargin]{footmisc} 

\title{Generalizable Lightweight Proxy for Robust NAS against Diverse Perturbations}

\author{%
 Hyeonjeong Ha$^{1\color{red}{*}}$, Minseon Kim$^{1}$\thanks{Equal contribution. Author ordering is determined by coin flip.}  , Sung Ju Hwang$^{1, 2}$\\
$^{1}$Korea Advanced Institute of Science and Technology (KAIST), $^{2}$DeepAuto\\
\texttt{\{hyeonjeongha, minseonkim, sjhwang82\}@kaist.ac.kr}  \\
}

\begin{document}
\maketitle

\begin{abstract}
Recent neural architecture search (NAS) frameworks have been successful in finding optimal architectures for given conditions (e.g., performance or latency). However, they search for optimal architectures in terms of their performance on clean images only, while robustness against various types of perturbations or corruptions is crucial in practice. Although there exist several robust NAS frameworks that tackle this issue by integrating adversarial training into one-shot NAS, however, they are limited in that they only consider robustness against adversarial attacks and require significant computational resources to discover optimal architectures for a single task, which makes them impractical in real-world scenarios. To address these challenges, we propose a novel lightweight robust zero-cost proxy that considers the consistency across features, parameters, and gradients of both clean and perturbed images at the initialization state. Our approach facilitates an efficient and rapid search for neural architectures capable of learning generalizable features that exhibit robustness across diverse perturbations. The experimental results demonstrate that our proxy can rapidly and efficiently search for neural architectures that are consistently robust against various perturbations on multiple benchmark datasets and diverse search spaces, largely outperforming existing clean zero-shot NAS and robust NAS with reduced search cost. Code is available at \href{https://github.com/HyeonjeongHa/CRoZe}{https://github.com/HyeonjeongHa/CRoZe}.
\end{abstract}
\section{Introduction}
Neural architecture search (NAS) techniques have achieved remarkable success in optimizing neural networks for given tasks and constraints, yielding networks that outperform handcrafted neural architectures~\citep{baker2016,liu2018progressive,luo2018neural,pham2018efficient,xu2019pc}. However, previous NAS approaches have primarily aimed to search for architectures with optimal performance and efficiency on clean inputs, while paying less attention to robustness against adversarial perturbations~\citep{goodfellow2014fgsm, madry2017pgd} or common types of corruptions~\citep{hendrycks2019benchmarking}. This can result in finding unsafe and vulnerable architectures with erroneous and high-confidence predictions on input examples even with small perturbations~\citep{mok2021advrush, jung2023neural}, limiting the practical deployment of NAS in real-world safety-critical applications.

\begin{wrapfigure}[12]{r}{0.56\textwidth}
    \vspace{-0.2in}
     \begin{subfigure}[b]{0.28\textwidth}
         \centering
         \includegraphics[height=3.0cm]{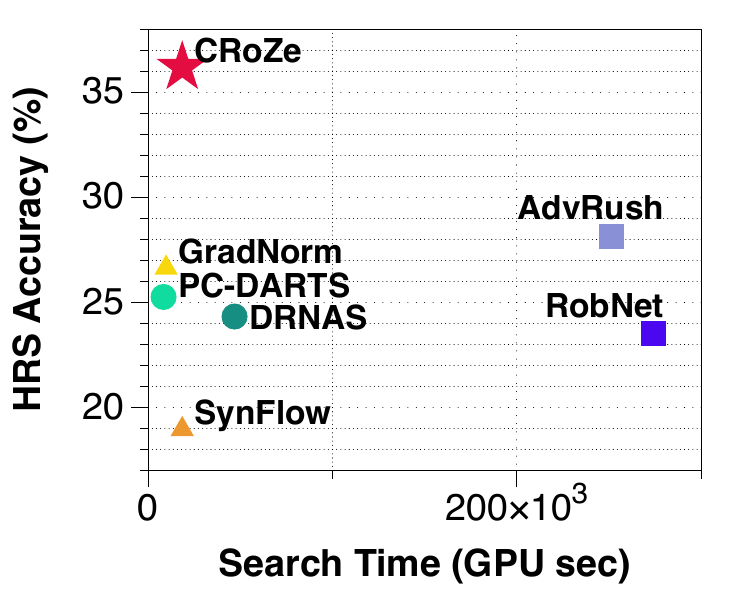}
         \vspace{-0.05in}
         \caption{\small{Standard-Trained}}
         \label{fig:e2e_st}
     \end{subfigure}
     \hspace{-0.1in}
     \begin{subfigure}[b]{0.28\textwidth}
         \centering
         \includegraphics[height=3.0cm]{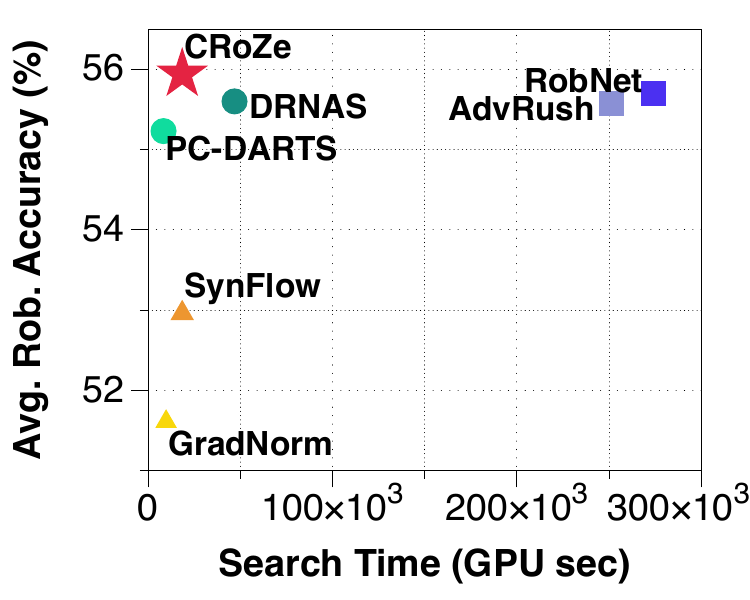}
         \vspace{-0.05in}
         \caption{\small{Adverserially-Trained}}
         \label{fig:e2e_at}
     \end{subfigure}
    \vspace{-0.05in}
    \caption{\small Final performance of the searched network in DARTS search space on CIFAR-10 through \textbf{{\textcolor{fig2green}{clean one-shot NAS}}, {\textcolor{fig2blue}{robust NAS}}, {\textcolor{fig2yellow}{clean zero-shot NAS}}} and \textbf{\textcolor{fig2red}{our CRoZe}}.}
    \label{fig:e2e_intro}
    \vspace{-0.05in}
\end{wrapfigure}

To address the gap between robustness and NAS, previous robust NAS works~\citep{mok2021advrush, guo2019meets} have proposed to search for adversarially robust architectures by integrating adversarial training into NAS. Yet, they are computationally inefficient as they utilize costly adversarial training on top of the one-shot NAS methods~\citep{liu2018darts, cai2019once}, requiring up to 33$\times$ larger computational cost than clean one-shot NAS~\citep{xu2019pc}. Especially,~\citet{guo2019meets} takes almost 4 GPU days on NVIDIA 3090 RTX GPU to train the supernet, as it requires performing adversarial training on subnets with perturbed examples (Figure~\ref{fig:e2e_intro}, RobNet). Furthermore, they only target a single type of perturbation, i.e., adversarial perturbation~\citep{goodfellow2014fgsm, madry2017pgd}, thus, failing to generalize to diverse perturbations. In order to deploy NAS to real-world applications that require handling diverse types of tasks and perturbations, we need a lightweight NAS method that can yield robust architectures without going over such costly processes.


\begin{figure}
    \centering
    \begin{subfigure}[t]{0.50\textwidth}
         \centering
         \includegraphics[width=\textwidth]{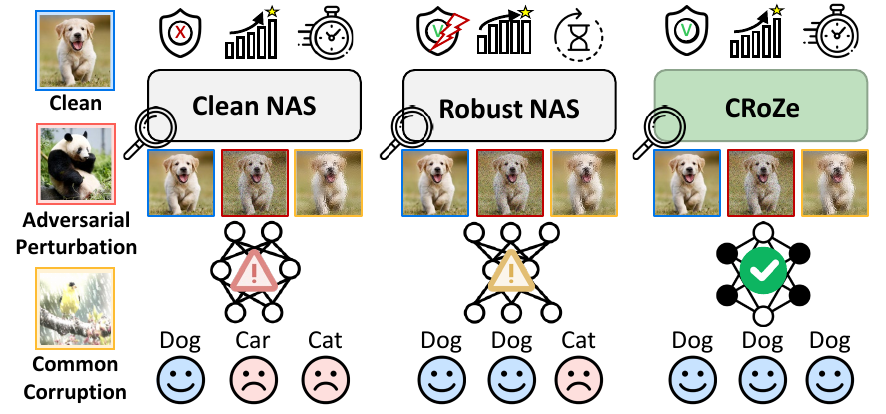}
         \vspace{-0.15in}
         \caption{\small Clean NAS and Robust NAS vs CRoZe.}
         \label{fig:concept2}
     \end{subfigure}
     \hfill
     \begin{subfigure}[t]{0.46\textwidth}
         \centering
         \includegraphics[width=0.95\textwidth]{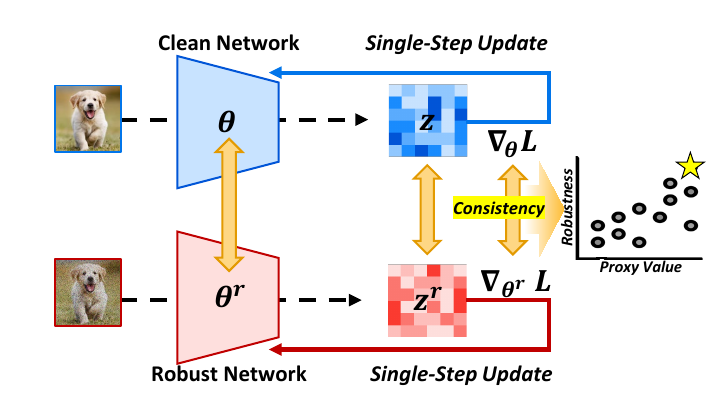}
         \caption{\small Consistency-based zero-cost proxy (CRoZe).}
         \label{fig:method}
     \end{subfigure}
    \caption{\small \textbf{Generalizable lightweight proxy for robust NAS against diverse perturbations.} While previous NAS methods search for neural architectures primarily on clean samples (Clean NAS) or adversarial perturbations (Robust NAS) with excessive search costs and fail to generalize across diverse perturbations, our proposed proxy, namely CRoZe, can rapidly search for high-performing neural architectures against diverse perturbations. Specifically, CRoZe evaluates the network's robustness in a single step based on the consistency across the features ($z$ and $z^r$), parameters ($\theta$ and $\theta^r$), and gradients ($\nabla_\theta \mathcal{L}$ and $\nabla_{\theta^r} \mathcal{L}$) between clean and robust network against clean and perturbed inputs, respectively.}
    \label{fig:concept}
\vspace{-0.2in}
\end{figure}

To tackle this challenge, we propose a novel and lightweight  \textbf{C}onsistency-based \textbf{Ro}bust \textbf{Ze}ro-cost proxy (\textbf{CRoZe}) that can rapidly evaluate the robustness of the neural architectures against \textit{diverse semantic-preserving} perturbations without requiring any iterative training. While prior clean zero-shot NAS methods~\citep{abdelfattah2021zero, mellor2021neural} introduced proxies that score the networks with randomly initialized parameters~\citep{lee2018snip, wang2020grasp, liu2021fisher, tanaka2020synflow} without any training, they only consider which parameters are highly sensitive to clean inputs for a given task, as determined by measuring the scale of the gradients based on the objectives and thus yield networks that are vulnerable against perturbed inputs (Figure~\ref{fig:concept2}).

Specifically, our proxy captures the consistency across the features, parameters, and gradients of a randomly initialized model for both clean and perturbed inputs, which is updated with a single gradient step (Figure~\ref{fig:method}). This metric design measures the model's robustness in multiple aspects, which is indicative of its generalized robustness to diverse types of perturbations. This prevents the metric from being biased toward a specific type of perturbation and ensures its robustness across diverse semantic-preserving perturbations. Empirically, we find that a neural architecture with the highest performance for a single type of perturbation tends to exhibit larger feature variance for other types of perturbations (Figure~\ref{fig:feat_dist}), while our proxy that considers the robustness in multiple aspects obtains features with smaller variance even on diverse types of perturbations. This suggests that our proxy is able to effectively discover architectures with enhanced generalized robustness.

We validate our approach through extensive experiments on diverse search spaces (NAS-Bench 201, DARTS) and multiple datasets (CIFAR-10, CIFAR-100, ImageNet16-120), with not only the adversarial perturbations but also with various types of common corruptions~\citep{hendrycks2019benchmarking}, against both clean zero-shot NAS~\citep{abdelfattah2021zero, mellor2021neural} and robust NAS baselines~\citep{mok2021advrush, guo2019meets}. The experimental results clearly demonstrate our method's effectiveness in finding generalizable robust neural architectures, as well as its computational efficiency. Our contributions can be summarized as follows:

\begin{itemize}
    \item We propose a simple yet effective consistency-based zero-cost proxy for robust NAS against diverse perturbations via measuring the consistency of features, parameters, and gradients between perturbed and clean samples.
    
    \item Our approach can rapidly search for generalizable neural architectures that do not perform well only on clean samples but also are highly robust against diverse types of perturbations on various datasets. 
    
    \item Our proxy obtains superior Spearman's $\rho$ across benchmarks compared to existing clean zero-shot NAS methods and identifies robust architectures that exceed robust NAS frameworks by \textbf{5.57\%} with \textbf{14.7 times less search cost} within the DARTS search space on CIFAR-10.
    
\end{itemize}

\section{Related Work}
\paragraph{Robustness of DNNs against Perturbations.}
Despite the recent advances, deep neural networks (DNNs) are still vulnerable to small perturbations on the input, e.g., common corruptions~\citep{hendrycks2019benchmarking}, random noises~\citep{dodge2017study}, and adversarial perturbations~\citep{biggio2013evasion,szegedy2013intriguing}, which can result in incorrect predictions with high confidence. To overcome such vulnerability against diverse perturbations, many approaches have been proposed to train the neural network to be robust against each type of perturbation individually. To learn a rich visual representation from limited crawled data, previous works~\citep{hendrycks2019augmix, cubuk2020randaugment} utilized a combination of strong data augmentation functions to improve robustness to common corruptions and random Gaussian noises. Furthermore, to overcome adversarial vulnerability, widely used defense mechanisms~\citep{goodfellow2014fgsm,madry2017pgd,moosavi2016deepfool} generate adversarially perturbed images by taking multiple gradient steps to maximize the training loss and use them in training to improve the model's robustness.

\vspace{-0.05in}
\paragraph{Neural Architecture Search.}
Neural architecture search (NAS) leverages reinforcement learning~\citep{zoph2016neural, baker2016, zhong2018practical} or evolutionary algorithms~\citep{real2017large, liu2017hierarchical, elsken2018efficient, real2019regularized} to automate the design of optimal architectures for specific tasks or devices. However, those are computationally intensive, making them impractical to be applied in real-world applications. To address this, zero-shot NAS methods~\citep{abdelfattah2021zero, mellor2021neural} have emerged that significantly reduce search costs by predicting the performance of architecture at the initialization state only with a single batch of a given dataset. Despite the improvement in NAS, previous zero-shot NAS methods, and conventional NAS methods aim only to find architectures with high accuracy on clean examples, without considering their robustness against various perturbations. In particular, SynFlow~\citep{abdelfattah2021zero} lacks data incorporation in its scoring mechanism, potentially failing to find the network that can handle diverse perturbations. NASWOT~\citep{mellor2021neural} search models with low activation correlation across two different inputs, emphasizing their distinguishability. However, the criterion contradicts the requirements of robustness, which necessitates finding a network that maintains similar activations for both clean and perturbed inputs. As a result, models found with previous NAS methods often lead to incorrect predictions with high confidence~\citep{mok2021advrush, jung2023neural} even with small imperceptible perturbations applied to the inputs. A new class of NAS methods~\citep{guo2019meets, mok2021advrush} that considers robustness against adversarial perturbations has emerged. Yet, they require adversarial training of the supernet, which demands more computational resources than conventional NAS~\citep{real2017large,elsken2018efficient} due to repeated adversarial gradient steps. Furthermore, adversarial robust NAS often overfit to a single type of perturbation due to only considering the adversarial robustness. Thus, there is a need for a lightweight NAS approach that can achieve generalized robustness for safe real-world applications.  

\section{Methods}
Our ultimate goal is to efficiently search for robust architectures that have high performance on various tasks, regardless of the type of perturbations applied to the input samples. To achieve this goal, we propose a \textbf{C}onsistency-based \textbf{Ro}bust \textbf{Ze}ro-cost proxy (\textbf{CRoZe})  that considers the consistency of the features, parameters, and gradients between a single batch of clean and perturbed inputs obtained by taking a single gradient step. The proposed proxy enables the rapid evaluation of the robustness of the neural architectures in the random states, without any adversarial training (Figure \ref{fig:concept}).

\subsection{Robust Architectures}
\label{sec:rob_archs}
Formally, our goal is to accurately estimate the final robustness of a given neural architecture $\mathcal{A}$ and a single batch of inputs $B=\{(x, y)\}$, without training. Here, $x \in X$ is the input sample, and $y \in Y$ is its corresponding label for the given dataset $\mathcal{D}$. In the following section, the network $f_{\theta}(\cdot)$ consists of an encoder $e_{\psi}(\cdot)$ and a linear layer $h_{\pi}(\cdot)$, which is $\mathcal{A}$ that is parameterized with $\psi$ and $\pi$, respectively. The straightforward approach to evaluating the robustness of the network is measuring the accuracy against the perturbed input $x'$ with unseen semantic-preserving perturbations, as follows:
\begin{equation}
    \text{Accuracy} = \frac{1}{N}\sum_{n=1}^N\mathbf{\zeta}(\argmax_{c\in Y}\mathbb{P}(h_{\pi}\circ e_{\psi}(x')=c)=y), \\
\end{equation}
where $\zeta$ is the Kronecker delta function, which returns 1 if the predicted class $c$ is the same as $y$, and 0 otherwise, $x'$ is a perturbed input, such as one with random Gaussian noise, common types of corruptions~\citep{hendrycks2019augmix}, or adversarial perturbations~\citep{madry2017pgd,goodfellow2014fgsm} applied to it. Specifically, to have a correct prediction on the unseen perturbed input $x'$, the model needs to extract similar features between $x'$ and $x$, assuming that the model can correctly predict the label for input $x$ as follows:
\begin{equation}
\label{eq:feature_bound}
    \parallel e_{\psi}(x) - e_{\psi}(x') \parallel \leq \epsilon, \\
\end{equation}
where $\epsilon$ is sufficiently small bound. Thus, a robust model is one that can extract consistent features across a wide range of perturbations. However, precisely assessing the accuracy of the model against perturbed inputs requires training from scratch with a full dataset, which incurs a linear increase in the computational cost with respect to the number of neural architectures to be evaluated.

\subsection{Estimating Robust Network through Perturbation}
In this section, we explain details on preliminary protocols before computing our proxy. Due to the impractical computation cost to obtain a combinatorial number of fully-trained models in a given neural architecture search space (i.e., $10^{19}$ for DARTS), we propose to utilize two surrogate networks which together can estimate the robustness of fully-trained networks within a single gradient step. The two surrogate networks are a clean network $f_{\theta}$ with the randomly initialized parameter $\theta$ and a robust network $f_{\theta^r}$ with the robust parameter $\theta^r$, which is determined with a parameter perturbations from $f_{\theta}$. Then, the obtained $\theta^r$ is used to generate the single batch of perturbed inputs for our proxy.

\vspace{-0.05in}
\paragraph{Robust Parameter Update via Layer-wise Parameter Perturbation.}
We employ a surrogate robust network to estimate the output of fully-trained networks against perturbed inputs. To make the perturbation stronger, we use a double-perturbation scheme that combines layer-wise parameter perturbations~\citep{awp20wu} and input perturbations, both of which maximize the training objectives $\mathcal{L}$. This layer-wise perturbation allows us to learn smoother updated parameters by min-max optimization, through which we can obtain the model with the maximal possible generalization capability~\citep{awp20wu,foret2021sharpnessaware} within a single step. Specifically, given a network $f$ is composed of $M$ layers, $f_\theta = f_{\theta_M} \circ \cdots \circ f_{\theta_1}$, with parameters $\theta = \{\theta_1, \dots, \theta_M\}$, the $m^{th}$ layer-wise parameter perturbation is done as follows:
\begin{equation}
\begin{aligned}
     \label{equation:attack_params}
    \theta^r_m \leftarrow \theta_m + \beta*\frac{\nabla_{\theta_m}\mathcal{L}\big(f_{\theta}(x), y\big)}{\lVert 
 \nabla_{\theta_m}\mathcal{L}\big(f_{\theta}(x), y\big)\rVert}* \lVert \theta_m\rVert,
\end{aligned}
\end{equation}
where $\beta$ is the step size for parameter perturbations, $\|\cdot\|$ is the norm, and $\mathcal{L}$ is the cross-entropy objective. This bounds the size of the perturbation by the norm of the original parameter $\|\theta_m\|$.

\vspace{-0.05in}
\paragraph{Perturbed Input.}
On top of the perturbed parameters (Eq.~\ref{equation:attack_params}), we generate perturbed input images by employing fast gradient sign method (FGSM)~\citep{goodfellow2014fgsm}, which is the worst case adversarial perturbation to the input $x$ as follows:
\vspace{-0.05in}
\begin{equation}
 \label{equation:attack_input}
     \small\delta = \epsilon \mathtt{sign}\Big(\nabla_{x} \mathcal{L}\big(f_{\theta^r}(x), y\big)\Big),
\end{equation}
where $\delta$ is a generated adversarial perturbation that maximizes the cross-entropy objective $\mathcal{L}$ of given input $x$ and given label $y$. Then, we utilize the perturbed inputs ($x'=x+\delta$) to estimate the robustness of the fully-trained model. Although CRoZe is an input perturbation-agnostic proxy (Supplementary~\ref{sec:ablation_input_perturbation}), we employ adversarially perturbed inputs for all the following sections.

\subsection{Consistency-based Proxy for Robust NAS}
\label{sec:method_croze}
We now elaborate the details on our proxy that evaluate the robustness of the architecture with the two surrogate networks: the clean network that is randomly initialized and uses clean images $x$ as inputs, and the robust network parameterized with $\theta^r$ which uses perturbed images $x+\delta$ as inputs. 

\vspace{-0.05in}
\paragraph{Features, Parameters, and Gradients.} As we described in Section~\ref{sec:rob_archs}, we first evaluate the representational consistency between clean input ($x$) and perturbed input ($x'$) by forwarding them through the encoder of clean surrogate network $f_\theta(\cdot)$ and robust surrogate network $f_{\theta^r}(\cdot)$, respectively, as follows:
\begin{align}
    \label{equation:feature}
    \mathcal{Z}_m(f_\theta(x), f_{\theta^r}(x')) = 1+\frac{z_m \cdot z_m^{r}}{\lVert z_m\rVert \lVert z_m^r\rVert}, 
\end{align}
where $z_m$ and $z_m^{r}$ are output feature of each network $f_\theta(\cdot)$ and $f_{\theta^r}(\cdot)$, respectively, from each $m^{th}$ layer. Especially, we measure layer-wise consistency with cosine similarity function between clean and robust features. The higher feature consistency infers the higher robustness of the network.

However, the proxy solely considering the feature consistency within a single batch can be heavily reliant on the selection of the batch.  Therefore, to complement the feature consistency, we propose incorporating the consistency of updated parameters and gradient conflicts from each surrogate network as additional measures to evaluate the robustness of the network. To introduce these concepts, let us first denote the gradient and updated parameter of each surrogate network. The gradient $g$ from the clean surrogate network $f_{\theta}$ and robust surrogate network $f_{\theta^r}$ against clean images $x$ and perturbed images $x'$, respectively, are obtained as follows:
\begin{equation}
\begin{aligned}
    \label{equation:gradient}
     g = \nabla_{\theta}\mathcal{L}\big(f_{\theta}(x), y),&&g^r = \nabla_{\theta^r}\mathcal{L}\big(f_{\theta^r}(x'), y),
\end{aligned}
\end{equation}
where $g$ and $g^r$ are the gradients with respect to cross-entropy objectives $\mathcal{L}$ for clean images $x$ and perturbed images $x'$, respectively. Then, we can acquire single-step updated clean parameters $\theta$ and robust parameters $\theta^r$ calculated with gradients $g$ and $g^r$ and learning rate $\gamma$, respectively as follows:
\begin{equation}
\begin{aligned}
 \label{equation:weight}
    \theta_1 &\leftarrow \theta - \gamma g, &&
    \theta_1^r &\leftarrow \theta^r - \gamma g^r.
\end{aligned}
\end{equation}
Since each surrogate network represents the model for each task, i.e., clean classification and perturbed classification, the parameters and gradients of each surrogate network correspond to the updated weights and convergence directions for each task. Thus, the network that has high robustness will exhibit identical or similar parameter spaces for both classification tasks. However, as acquiring parameters of a fully-trained network is impractical, we estimate the converged parameters with the single-step updated parameters $\theta_1$ and $\theta^r_1$. Accordingly, since the higher similarity of single-step updated parameters may promote the model to converge to an identical or similar parameter space for both tasks, we evaluate the parameter similarity as one of our proxy terms as follows:
\begin{align}
    \mathcal{P}_m(\theta_1, \theta_1^r) = 1 + \frac{\theta_{1,m}\cdot \theta_{1,m}^r}{\lVert \theta_{1,m}\rVert \lVert \theta_{1,m}^r\rVert}.
\vspace{-0.1in}
\end{align}

Moreover, each gradient of the surrogate networks represents the converged direction of given objectives for each task, which is cross-entropy loss of clean input and perturbed input (Eq.~\ref{equation:gradient}). We employ the similarity of these gradients to assess the difficulties of optimizing architecture for both tasks. When the gradient directions are highly aligned between the two tasks, the learning trajectory for both tasks becomes more predictable, facilitating the optimization of both tasks easily. In contrast, orthogonal gradient directions lead to greater unpredictability, hindering optimization and potentially resulting in suboptimality for both clean or perturbed classification tasks. Therefore, to evaluate the stability of optimizing both tasks, we measure the absolute value of gradient similarity as follows:
\begin{align}
    \mathcal{G}_m(g, g^r) = \Bigg\lvert\frac{g_m \cdot g_m^{r}}{\lVert g_m\rVert \lVert g_m^r\rVert}\Bigg\rvert.
\end{align}

\vspace{-0.1in}
\paragraph{Consistency-based Robust Zero-Cost Proxy: CRoZe.}
In sum, to evaluate the robustness of the given architecture, we propose a scoring mechanism that evaluates the consistencies of features, parameters, and gradients between the clean network $f_\theta$ and the robust network $f_{\theta^r}$ that are obtained with a single gradient update. The robustness score for a given architecture is computed as follows:
\vspace{-0.1in}
\begin{equation}
 \label{equation:proxy}
     \text{CRoZe}(x, x';f_\theta, f_{\theta^r}) = \sum_{m=1}^{M} \mathcal{Z}_m \times \mathcal{P}_m \times \mathcal{G}_m.
\end{equation}
That is, we score the network $f_\theta$ with a higher CRoZe score as more robust to perturbations. In the next section, we show that this measure is highly correlated with the actual robustness of a fully-trained model (Table \ref{tbl:src_nasbench201_fastat}).


\section{Experiments \label{sec:experiments}}
We now experimentally validate our proxy designed to identify robust architectures that perform well on both clean and perturbed inputs, on multiple benchmarks. We first evaluate Spearman's $\rho$ between robustness and the proxy's values across different tasks and perturbations in the NAS-Bench-201 search space, comparing it to clean zero-shot NAS methods (Section~\ref{sec:nasbench201_clean}). We then evaluate the computational efficiency and final performance of the chosen architecture using our proxy in the DARTS search space, comparing it to existing robust NAS methods, which are computationally costly (Section~\ref{sec:e2e_darts}). Lastly, we analyze the proxy's ability to accurately reflect the fully trained model's behavior in a single step, as well as the capacity of the chosen robust architecture to consistently generate features, aligned gradients, and parameters against clean and perturbed inputs (Section~\ref{sec:analysis}).

\subsection{Experimental Setting}

\paragraph{Baselines.}
We consider three types of existing NAS approaches as our baselines. \textbf{1) Clean one-shot NAS}~\citep{xu2019pc, chen2020drnas} One-shot NAS methods for searching architectures only on clean samples. \textbf{2) Clean zero-shot NAS}~\citep{abdelfattah2021zero, mellor2021neural}: Zero-shot NAS with proxies that evaluate the clean performance of architectures without any training.  \textbf{3) Robust NAS}~\citep{guo2019meets, mok2021advrush} One-shot NAS methods for searching architectures only on adversarially perturbed samples.

\vspace{-0.1in}
\paragraph{Datasets.} 
For the NAS-Bench-201~\citep{nasbench201, jung2023neural} search space, we validate our proxy across different tasks (CIFAR-10, CIFAR-100, and ImageNet16-120) and perturbations (FGSM~\citep{goodfellow2014fgsm}, PGD~\citep{madry2017pgd}, and 15 types of common corruptions~\citep{hendrycks2019benchmarking}). To measure Spearman's $\rho$ between final accuracies and our proxy values, we use both clean NAS-Bench-201~\citep{nasbench201} and robust NAS-Bench-201~\citep{jung2023neural}, which include clean accuracies and robust accuracies. Finally, we search for the optimal architectures with our proxy in DARTS~\citep{liu2018darts} search space and compare the final accuracies against previous NAS methods~\citep{mok2021advrush, guo2019meets, chen2020drnas, abdelfattah2021zero} on CIFAR-10 and CIFAR-100.

\vspace{-0.1in}
\paragraph{Standard Training \& Adversarial Training.}
For a fair comparison, we use the same training and testing settings to evaluate all the architectures searched with all NAS methods, including ours. 1) Standard Training: We train the neural architectures for 200 epochs under SGD optimizer with a learning rate of 0.1 and weight decay 1e-4, and use a batch size of 64 following~\citep{mok2021advrush}. 2) Adversarial Training: We train the neural architectures with $l_\infty$ PGD attacks with the epsilon of 8/255, step size 2/255, and 7 steps. We evaluate the robustness against various perturbations, which are adversarial attacks (FGSM~\cite{goodfellow2014fgsm}, PGD~\cite{madry2017pgd}, CW~\cite{carlini2017cw}, DeepFool~\cite{moosavi2016deepfool}, SPSA~\cite{uesato2018adversarial}, LGV~\cite{gubri2022lgv}, and AutoAttack~\cite{croce2020reliable}) and common corruptions~\cite{hendrycks2019benchmarking}. 
More experimental details are described in Supplementary~\ref{sup:exp_setting}.

\subsection{Results on NAS-Bench-201}

\paragraph{Standard-Trained Neural Architectures.}
\label{sec:nasbench201_clean}
\begin{table}[t]
\caption{\small Comparison of Spearman's $\rho$ between the actual accuracies and the proxy values on CIFAR-10 in the NAS-Bench-201 search space. Clean stands for clean accuracy and robust accuracies are evaluated against diverse adversarial attacks~\cite{goodfellow2014fgsm, carlini2017cw, moosavi2016deepfool, uesato2018adversarial, gubri2022lgv, croce2020reliable}. * denotes the HRS value of each attack and AA. indicates AutoAttack. Avg. stands for average Spearman's $\rho$ values with all accuracies. \textbf{Bold} and \underline{underline} stands for the best and second. All models are trained on adversarially-perturbed images.}
    \begin{center}
    \vspace{-0.05in}
        \resizebox{0.95\linewidth}{!}{
        \begin{tabular}{lcccccccccca}
            \toprule
            Proxy Type & Clean & FGSM & PGD & FGSM* & PGD* & CW & DeepFool & SPSA & LGV & AA. & Avg. \\
            \midrule
            \midrule
            FLOPs & 0.670 & 0.330 & 0.418 & 0.531 & 0.515 & 0.189 & 0.364 & 0.196 & 0.347 & 0.365 & 0.393\\
            \#Params. & \underline{0.678} & 0.341 & \underline{0.429} & \underline{0.541} & \underline{0.526} & 0.182 & 0.371 & 0.209 & 0.355 & 0.375 & \underline{0.401} \\
            Plain~\cite{abdelfattah2021zero} & -0.042 & -0.007 & -0.012 & -0.016 & -0.016 & 0.072 & 0.009 & 0.009 & 0.010 & 0.015 & 0.002 \\
            Grasp~\cite{abdelfattah2021zero} & 0.470 & 0.324 & 0.341 & 0.392 & 0.375 & 0.179 & 0.393 & \textbf{0.249} & \underline{0.401} & 0.397 & 0.352\\
            Fisher~\cite{abdelfattah2021zero} & 0.482 & 0.226 & 0.276 & 0.335 & 0.334 & 0.234 & 0.242 & 0.092 & 0.239 & 0.244 & 0.270 \\
            GradNorm~\cite{abdelfattah2021zero} & 0.659 & 0.336 & 0.400 & 0.490 & 0.478 & \textbf{0.264} & \underline{0.421} & 0.149 & \underline{0.401} & \underline{0.405} & 0.400 \\
            SynFlow~\cite{abdelfattah2021zero} & 0.635 & \underline{0.355} & 0.420 & 0.519 & 0.498 & 0.202 & 0.397 & 0.196 &0.387& 0.383 & 0.399 \\
            NASWOT~\cite{mellor2021neural} & 0.600 & 0.332 & 0.381 & 0.437 & 0.438 & \underline{0.240} & 0.250 & 0.197 &0.265& 0.280 & 0.342 \\
            \midrule
            CRoZe & \textbf{0.723} & \textbf{0.417} & \textbf{0.501} & \textbf{0.602} &\textbf{0.588} & 0.220  & \textbf{0.454} & \underline{0.240} &\textbf{0.449}& \textbf{0.458} & \textbf{0.465} \\
            \midrule
            \bottomrule
            \end{tabular} }
            \vspace{-0.1in}
            \label{tbl:src_nasbench201_fastat}
    \vspace{-0.18in}
    \end{center}
\end{table}

\begin{table}[t]
    \begin{center}
    \caption{\small Comparison of Spearman's $\rho$ between the actual accuracies and the proxy values on CIFAR-10, CIFAR-100 and ImageNet16-120 in NAS-Bench-201 search space. Clean stands for clean accuracy and robust accuracies are evaluated against adversarial perturbations (FGSM)~\cite{goodfellow2014fgsm} with various attack sizes ($\epsilon$) and common corruptions~\citep{hendrycks2019benchmarking}. Avg. stands for average Spearman's $\rho$ values with all accuracies within each task. \textbf{Bold} and \underline{underline} stands for the best and second. All models are standard-trained on clean images.}
    
    \label{tbl:src_nasbench201_clean_all}
    \resizebox{1.0\linewidth}{!}{
        \small {
        \begin{tabular}{lcccccccaccccccacca}
            \toprule
            \centering
             \multirow{3}{*}{Proxy Type} & \multicolumn{8}{c}{CIFAR-10} & \multicolumn{7}{c}{CIFAR-100} & \multicolumn{3}{c}{ImageNet16-120} \\
            \cmidrule(r){2-9}\cmidrule(r){10-16}\cmidrule(r){17-19}
             & \multirow{2}{*}{Clean} & \multicolumn{2}{c}{FGSM} & \multicolumn{4}{c}{Common Corruption (CC.)} & \cellcolor{white} & \multirow{2}{*}{Clean} & FGSM & \multicolumn{4}{c}{Common Corruption (CC.)} & \cellcolor{white} & \multirow{2}{*}{Clean} & FGSM & \cellcolor{white} \\
             \cmidrule(r){3-4}\cmidrule(r){5-8}\cmidrule(r){11-11}\cmidrule(r){12-15}\cmidrule(r){18-18} 
             & & $\epsilon=8$ & $\epsilon=4$ & Weather & Noise & Blur & Digital & Avg. & & $\epsilon=4$ & Weather & Noise & Blur & Digital & Avg. & & $\epsilon=4$ & Avg. \\ 
            \midrule
            \midrule
             FLOPs & 0.726 & 0.753 & 0.740 & 0.665 & 0.138 & \underline{0.232} &  0.473 & 0.532 & 0.699 & 0.661 & 0.674  & 0.253 & 0.492 &  0.607 & 0.557 & 0.680 & 0.611 & 0.634 \\
            \#Params. & 0.747 & 0.756 & 0.739 & \underline{0.674} & 0.131 & 0.229 & 0.489 & 0.502 & 0.720 & 0.654 & 0.685 & 0.240 & 0.489 &  0.618 & 0.568 & 0.683 & 0.627 & 0.655 \\
            Plain~\cite{abdelfattah2021zero} & -0.073 & -0.059 & -0.055 & -0.041 & 0.035 & 0.048 &  -0.032 & -0.025 & -0.242 & -0.172 & -0.169 & 0.055 & -0.110 &  -0.182 & -0.137 & -0.231 & -0.241 & -0.236 \\
            Grasp~\cite{abdelfattah2021zero} & 0.551 & 0.657 & 0.673 & 0.641 & \textbf{0.250} & 0.226 & 0.426 & 0.532 & 0.550 & 0.649 & 0.565 & 0.311 & 0.468 &  0.518 & 0.510 & 0.543 & 0.601 & 0.572 \\
            Fisher~\cite{abdelfattah2021zero} & 0.383 & 0.466 & 0.499 & 0.411 & 0.179 & 0.221 &  0.244 & 0.343 & 0.382 & 0.575 & 0.425 & \underline{0.325} & 0.447 &  0.396 & 0.425 & 0.324 & 0.394 & 0.359 \\
            GradNorm~\cite{abdelfattah2021zero} & 0.637 & 0.760 & \underline{0.774} & 0.540 & \underline{0.239} & 0.183 &  0.375 & 0.460 & 0.638 & \textbf{0.789} & 0.654 & \textbf{0.357} & \textbf{0.562} &  0.613 & \underline{0.602} & 0.573 & 0.649 & 0.611 \\
            SynFlow~\cite{abdelfattah2021zero} & \underline{0.777} & \underline{0.778} & 
            0.751 & 0.673 & 0.188 & 0.181 &  \underline{0.554} & \underline{0.557} & \underline{0.769} & 0.683 & \underline{0.703} & 0.215 & 0.439 &  \underline{0.641} & 0.575 & \underline{0.751} & 0.695 & \underline{0.723} \\
            NASWOT~\cite{mellor2021neural} & 0.687 & 0.522 & 0.513 & 0.479 & -0.078 & 0.012 &  0.434 & 0.367 &  0.708 & 0.520 & 0.525 & -0.023 & 0.319 &  0.534 & 0.431 & 0.698 & \textbf{0.730} & 0.714 \\
            \midrule
            CRoZe & \textbf{0.823} & \textbf{0.826} & \textbf{0.801} & \textbf{0.743} & 0.190 & \textbf{0.236} &  \textbf{0.565} & \textbf{0.598} & \textbf{0.784} & \underline{0.693} & \textbf{0.747} & 0.251 & \underline{0.504} &  \textbf{0.682} & \textbf{0.610} & \textbf{0.765} & \underline{0.696} & \textbf{0.731} \\
            \midrule
            \bottomrule
            \end{tabular} }}
            \vspace{-0.2in}
    \end{center}
\end{table}

In order to verify the effectiveness of our proxy in searching for high-performing neural architectures across various tasks and perturbations, we conduct experiments using Spearman's $\rho$ as a metric to evaluate the preservation of the rank between the proxy values and final accuracies~\citep{abdelfattah2021zero, mellor2021neural, dong2023diswot}. For Spearman's $\rho$ between clean accuracies and proxy values, existing clean zero-shot NAS works~\citep{abdelfattah2021zero, mellor2021neural} performed worse than using the number of parameters as a proxy (\#Params.). In contrast, our proxy shows significantly higher correlations with clean accuracies across all tasks, demonstrating improvements of 5.92\% and 1.95\% on CIFAR-10 and CIFAR-100, respectively, compared to best-performing baselines (Table~\ref{tbl:src_nasbench201_clean_all}).

Furthermore, CRoZe shows remarkable Spearman's $\rho$ for robust accuracies obtained against adversarial perturbations and corrupted noises across tasks. Notably, our proxy outperforms the SynFlow~\citep{abdelfattah2021zero} by 6.65\% and 9.11\% in an average of Spearman's $\rho$ for adversarial perturbations and common corruptions on CIFAR-10, respectively (Table~\ref{tbl:src_nasbench201_clean_all}). Our results on multiple benchmark tasks with diverse perturbations highlight the ability of our proxy to effectively search for robust architectures that can consistently outperform predictions against various perturbations. Importantly, our proxy is designed to prioritize generalizability, and as a result, it exhibits consistently enhanced correlation with final accuracies for both clean and perturbed samples. This result indicates that considering generalization ability is effective in identifying robust neural architectures against diverse perturbations but also leads to improved performance for clean neural architectures.

\vspace{-0.05in}
\paragraph{Adversarially-Trained Neural Architectures.} 
\label{sec:nasbench201_adv}
We also validate the ability of our proxy to precisely predict the robustness of adversarially-trained networks, specifically for adversarial perturbations. Adversarial training~\citep{madry2017pgd} is a straightforward approach to achieving robustness in the presence of adversarial perturbations. To assess the Spearman's $\rho$ of robustness in adversarially-trained networks, we construct a dataset consisting of final robust accuracies of 500 randomly sampled neural architectures from NAS-Bench-201 search space that are adversarially-trained~\citep{madry2017pgd} from scratch. The final robust accuracies are obtained for 7 different adversarial attacks, including FGSM, PGD, CW, DeepFool, SPSA, LGV, and AutoAttack.

Notably, our proxy showcases the highest overall correlation for the 7 different adversarial attacks, supporting the ability of our proxy to search the architectures that are robust against diverse perturbations. Considering the trade-off between the clean and robust accuracy in adversarial training~\citep{zhang2019trades}, we employ the harmonic robustness score (HRS)~\citep{devaguptapu2020empirical} to evaluate the overall performance of the adversarially-trained models. When comparing the correlations between clean performances and proxy values, existing clean zero-shot NAS methods, i.e., SynFlow and Grasp, demonstrate higher correlations in the FGSM or PGD, but their correlations with clean accuracies are poorer than GradNorm and Fisher, respectively. This result shows that clean zero-shot NAS methods tend to search for architectures that are more prone to overfitting to either clean or robust tasks (Table~\ref{tbl:src_nasbench201_fastat}, \ref{tbl:src_nasbench201_clean_all}). In contrast, CRoZe consistently achieves higher Spearman's $\rho$ for both clean and robust tasks, ultimately enabling the search for architecture with high HRS due to consideration of alignment in gradients.

\begin{figure}
    \centering
    \begin{subfigure}[b]{0.49\textwidth}
         \begin{subfigure}[b]{0.49\textwidth}
             \centering
             \includegraphics[width=\textwidth]{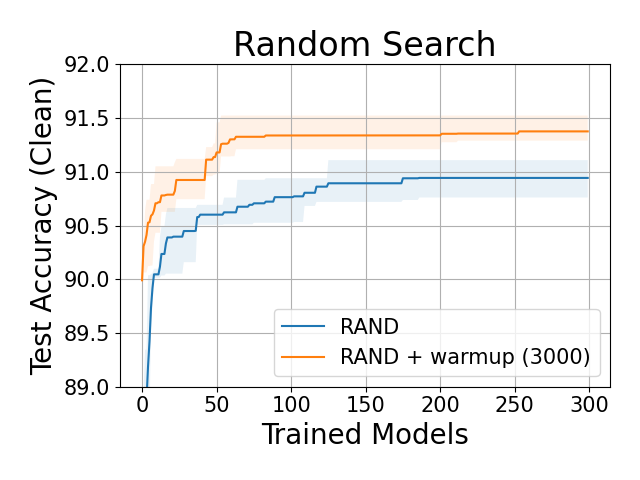}
             \vspace{-0.15in}
             \vspace{-0.1in}
             \label{fig:clean_random_search}
         \end{subfigure}
         \begin{subfigure}[b]{0.49\textwidth}
             \centering
             \includegraphics[width=\textwidth]{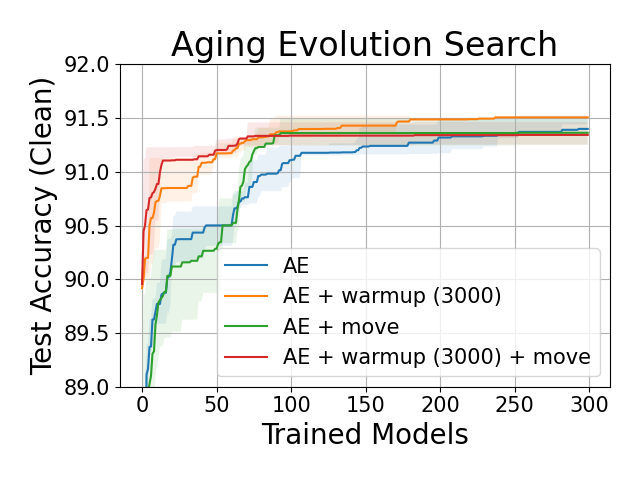}
             \vspace{-0.15in}
             \vspace{-0.1in}
             \label{fig:clean_ae}
         \end{subfigure}
         \caption{Search on Clean Inputs}
         \label{fig:warmup_clean}
    \end{subfigure}
     \begin{subfigure}[b]{0.49\textwidth}
         \begin{subfigure}[b]{0.49\textwidth}
             \centering
             \includegraphics[width=\textwidth]{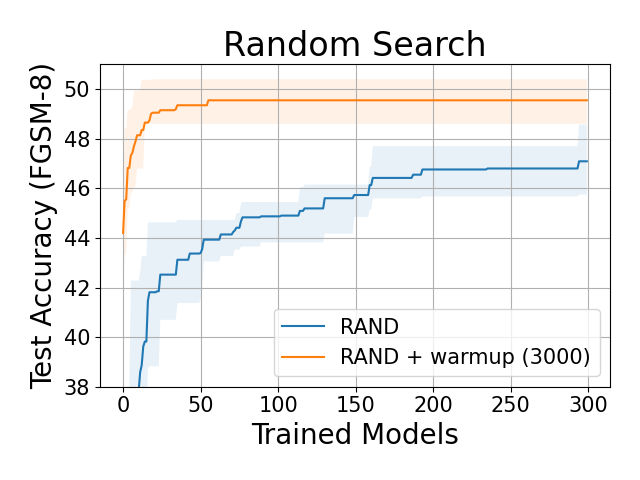}
             \vspace{-0.15in}
             \vspace{-0.1in}
             \label{fig:adv_random_search}
         \end{subfigure}
         \begin{subfigure}[b]{0.49\textwidth}
             \centering
             \includegraphics[width=\textwidth]{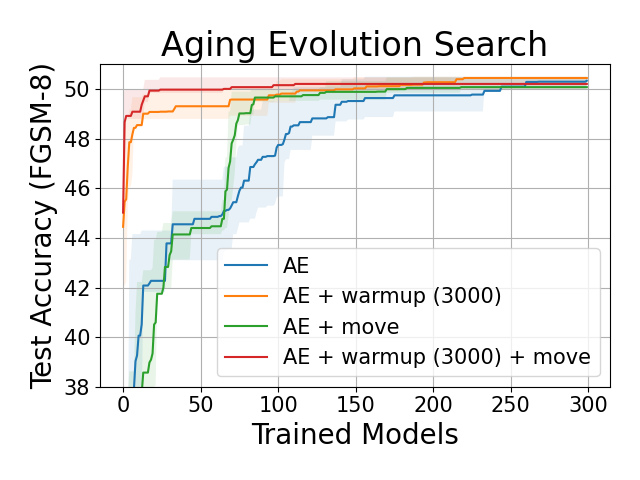}
             \vspace{-0.15in}
             \vspace{-0.1in}
             \label{fig:adv_ae_search}
         \end{subfigure}
         \caption{Search on Adversarially Perturbed Inputs}
         \label{fig:warmup_adv}
    \end{subfigure}
    \vspace{-0.08in}
    \caption{\small \textbf{Search with our proxy on CIFAR-10 in NAS-Bench-201 search space.} Our proxy can reduce the sampling costs in searching for neural architectures against both clean (left) and perturbed samples (right).}
    \label{fig:warmup}
    \vspace{-0.18in}
\end{figure}
\subsection{End-to-End Generalization Performance on DARTS}
\label{sec:e2e_darts}
In this section, we evaluate the effectiveness of CRoZe in rapidly searching for generalized neural architectures in the DARTS search space and compare it with previous clean one-shot NAS~\citep{xu2019pc, chen2020drnas}, clean zero-shot NAS~\citep{tanaka2020synflow}, and robust NAS~\citep{mok2021advrush, guo2019meets} in terms of performance and computational cost. 

\paragraph{Efficient Sampling with CRoZe.}

The efficiency of our proxy lies in its ability to evaluate neural architectures without iteratively training the models with those architectures. As a result, the majority of the search time is dedicated to sampling the candidate architectures. To thoroughly explore the sampling costs involved in using our proxy to identify good neural architectures, we conduct experiments in the NAS-Bench-201 search space with two representative sampling-based search algorithms: random search (RAND) and evolutionary search (AE). 

Our experiments show that CRoZe can rapidly identify an architecture with 91.5\% clean accuracy and 50\% robust accuracy against FGSM attacks, even at the initial stages of the search. By leveraging the initial pool of architectures with high proxy values obtained from CRoZe (RAND+warmup and AE+warmup), our proxy effectively reduces the sampling costs and rapidly identifies high-performing architectures for clean inputs (Figure~\ref{fig:warmup_clean}). Moreover, by focusing sampling around the architecture with the highest proxy values of CRoZe (AE+warmup+move),  we can search for architecture with even better robust performance using a less number of sampled models ($<50$) (Figure~\ref{fig:warmup_adv}). Overall, our results highlight the effectiveness of CRoZe in searching for high-performing neural architectures against both clean and perturbed inputs in the NAS-Bench-201 search space.

\vspace{-0.05in}
\paragraph{End-to-End Performance.}
We validate the final performance of the neural architectures discovered by CRoZe (Table~\ref{tbl:e2e},~\ref{tbl:e2e_adv}) and compare the search time and performance with existing NAS methods including robust one-shot NAS (RobNet~\citep{guo2019meets} and AdvRush~\citep{mok2021advrush}), clean one-shot NAS (PC-DARTS~\citep{xu2019pc} and DrNAS~\citep{chen2020drnas}), and clean zero-shot NAS (SynFlow, GradNorm~\citep{abdelfattah2021zero}). For a fair comparison between clean zero-shot NAS (SynFlow, GradNorm) and CRoZe, we sample the same number (e.g., 5,000) of candidate architectures using the warmup+move strategy in the DARTS search space. We report the average and standard deviation of accuracies evaluated over 3 different neural architecture searches. All experiments are conducted on a single NVIDIA 3090 RTX GPU to measure search costs.

\begin{table}[t]
\centering
\caption{\small Comparisons of the final performance and search time of the searched network in DARTS search space on CIFAR-10 and CIFAR-100. All NAS methods are executed on a single NVIDIA 3090 RTX GPU.}
\label{tbl:e2e}
\resizebox{\linewidth}{!}{
		\begin{tabular}{clcccccca}
	   	\toprule
	   	\multirow{2}{*}{Task}                          & 
	   	\multicolumn{1}{c}{\multirow{2}{*}{NAS Method}} & 
	   	Training-free                                    &
	   	Params                                      & 
	   	Search Time                                 & 
	   	\multicolumn{4}{c}{Standard-Trained}        \\
        \cmidrule{6-9}
	   	                                          & 
	   	                                          & 
	    NAS                                       & 
	   	(M)                                       & 
	   	(GPU sec)                                 &
            Clean                                       &
            CC.                                         &
            FGSM                                        &
            HRS                                         \\
	   	\midrule
	   	\midrule
        \multirow{7}{*}{CIFAR-10}                   &
	   	PC-DARTS~\citep{xu2019pc}                   &
	   	                                        &
	   	3.60                                        &
	   	8355                                        &
	   	\textbf{95.14} ($\pm$ 0.16)                              &
	   	73.57 ($\pm$ 0.89)	   		   	                        &
            17.74 ($\pm$ 3.44)                                      &
            29.76 ($\pm$ 4.81)                                       \\
	   	&
	   	DrNAS~\citep{chen2020drnas}                 &
	   	                                          &
	   	4.10                                        &
	   	46857                                       &
	    94.18 ($\pm$ 0.33)                                       &
	    71.79 ($\pm$ 0.74)	   		   	                        &
            15.63 ($\pm$ 2.78)                                       &
            26.71 ($\pm$ 4.01)                                      \\
            &
            RobNet~\citep{guo2019meets}            &
	   	                                        &
	   	5.44                                        &
	   	274062                                      &
	    94.94 ($\pm$ 0.26)                            &
	    72.49 ($\pm$ 0.45)	   		   	            &
            14.69 ($\pm$ 1.15)                            &
            25.42 ($\pm$ 1.71)                                       \\
            &
            AdvRush~\citep{mok2021advrush}              &
	   	                                        &
	   	4.20                                        &
	   	251245                                      &
	    93.44 ($\pm$ 0.41)                                       &
	    71.43 ($\pm$ 1.07)	   		   	                        &
            14.94 ($\pm$ 2.56)                                       &
            25.67 ($\pm$ 3.87)                                       \\
            &
            GradNorm~\citep{abdelfattah2021zero}         &
	   	\checkmark                                  &
	   	4.69                                        &
	   	9740                                        &
            92.95 ($\pm$ 0.67)                             &
            63.03 ($\pm$ 6.33)                             &
	   	15.64 ($\pm$ 0.15)                             &
	   	26.78 ($\pm$ 0.24)                             \\
	    &
            SynFlow~\citep{abdelfattah2021zero}         &
	   	\checkmark                                  &
	   	5.08                                        &
	   	10138                                       &
            92.15 ($\pm$ 1.25)                             &
            70.74 ($\pm$ 2.69)                             &
	   	12.65 ($\pm$ 1.46)                             &
	   	22.22 ($\pm$ 2.32)                             \\
            &
            CRoZe                                        &
	   	\checkmark                                  &
	   	5.52                                        &
	   	17066                                       &
	   	94.34 ($\pm$ 0.08)                             &
	   	\textbf{74.11} ($\pm$ 1.40) 		   	        &
            \textbf{20.51} ($\pm$ 1.32)                    &
            \textbf{33.07} ($\pm$ 2.30)                             \\
            \midrule
	   	\multirow{7}{*}{CIFAR-100}                  &
	   	PC-DARTS~\citep{xu2019pc}                   &
	   	                                        &
	   	3.60                                         &
	   	8355                                        &
	   	\textbf{76.95} ($\pm$ 0.67)                                       &
	   	\textbf{50.02} ($\pm$ 0.05)	   		   	                        &
            7.68 ($\pm$ 0.47)                                        &
            13.96 ($\pm$ 0.80)                                       \\
	   	&
	   	DrNAS~\citep{chen2020drnas}                 &
	   	                                          &
	   	4.10                                        &
	   	46857                                       &
	    75.49 ($\pm$ 1.66)                           &
	    49.18 ($\pm$ 1.31)	   		   	                &
            8.14 ($\pm$ 0.22)                                        &
            14.69 ($\pm$ 0.35)                                       \\
            &
            RobNet~\citep{guo2019meets}              &
	   	                                          &
	   	5.44                                         &
	   	274062                                       &
	    76.40 ($\pm$ 0.30)                                       &
	    49.77 ($\pm$ 0.82)                                       &
            7.50 ($\pm$ 0.84)                                        &
            13.65 ($\pm$ 1.39)                                       \\
            &
            AdvRush~\citep{mok2021advrush}              &
	   	                                          &
	   	4.20                                         &
	   	251245                                      &
	    74.48 ($\pm$ 1.35)                                       &
	    48.27 ($\pm$ 1.42)	   		   	                        &
            7.46 ($\pm$ 0.81)                                        &
            13.55 ($\pm$ 1.37)                                      \\
            &
            GradNorm~\citep{abdelfattah2021zero}         &
	   	\checkmark                                  &
	   	3.83                                        &
	   	9554                                        &
            68.52 ($\pm$ 0.80)                                      &
	   	43.69 ($\pm$ 2.22)                                       &
            6.87 ($\pm$ 1.58)                                        &
            12.44 ($\pm$ 2.61)                                       \\
	    &
            SynFlow~\citep{abdelfattah2021zero}         &
	   	\checkmark                                  &
	   	4.42                                        &
	   	9776                                        &
            76.23 ($\pm$ 0.73)                                       &
	   	49.60  ($\pm$ 0.78)                                     &
            9.01 ($\pm$ 0.56)                                        &
            16.11 ($\pm$ 0.91)                                       \\
            &
            CRoZe                                       &
	   	\checkmark                                 &
	   	4.72                                        &
	   	17457                                       &
	   	75.46 ($\pm$ 0.81)                                      &
	   	49.33 ($\pm$ 0.38)                                       &
            \textbf{9.75} ($\pm$ 0.78)                            &
            \textbf{17.26} ($\pm$ 1.22)                              \\
	   	\midrule
	   	\bottomrule
        \end{tabular}}
        \vspace{-0.24in}
\end{table}

\begin{wraptable}[10]{r}{0.5\textwidth}
    \centering
    \vspace{-0.18in}
    \caption{\small Final performance of the searched networks that are adversarially-trained on CIFAR-10.}
    \resizebox{\linewidth}{!}{
        \begin{tabular}{lcccccca}
        \toprule
        NAS                         &
              &
        \multicolumn{6}{c}{Robustness} \\
        \cmidrule{3-8}
        Method                      &
        Clean                       &
        PGD                         &
        CW                          &
        SPSA                        &
        LGV                         &
        AA.                         &
        Avg.                        \\
        \midrule
        PC-DARTS                    &
        86.02                       &
        52.16                       &
        9.85                        &
        86.00                       &
        79.32                       &
        48.82                       &
        55.23                      \\
        DrNAS                       &
        86.45                       &
        54.66                       &
        6.44                        &
        85.86                       &
        79.64                       &
        52.40                       &
        55.60                       \\
        RobNet                      &
        80.53                       &
        50.62                       &
        22.91                        &
        80.85                       &
        77.79                       &
        46.34                       &
        \underline{55.70}                       \\
        AdvRush                     &
        85.98                       &
        53.89                       &
        6.68                        &
        84.81                       &
        79.61                       &
        51.88                       &
        55.57                       \\
        GradNorm                    &
        81.61                       &
        49.86                       &
        12.02                        &
        77.18                       &
        73.27                       &
        46.69                       &
        51.61                       \\
        SynFlow                     &
        77.08                       &
        45.95                       &
        26.50                       &
        75.78                       &
        74.14                       &
        42.45                       &
        52.96                       \\
        CRoZe                       &
        84.28                       &
        52.17                       &
        19.13                        &
        83.43                       &
        76.85                       &
        48.14                       &
        \textbf{55.94}                       \\
        
        \bottomrule
        \end{tabular}}
    
    \label{tbl:e2e_adv}
    \vspace{-0.1in}
\end{wraptable}

In standard training, the searched network by our proxy surpasses the networks by robust one-shot NAS methods, RobNet and AdvRush, in terms of robust accuracy against FGSM on CIFAR-10, achieving improvements of 5.82\% and 5.57\%, respectively (Table~\ref{tbl:e2e}). Notably, CRoZe also shows the highest HRS accuracy with an 8.56\% increase compared to AdvRush on CIFAR-10, indicating that the neural architectures discovered by CRoZe effectively mitigate the trade-off between clean and robust accuracy, even with 14.7 times reduced search cost. When compared to the previously best-performing clean zero-cost proxy, SynFlow, CRoZe finds architectures with significantly superior performance across clean, common corruptions, and FGSM scenarios, showcasing the effectiveness of our proxy in identifying generalized architectures. Moreover, the neural architecture chosen by CRoZe outperforms clean one-shot NAS methods in HRS accuracy, effectively addressing vulnerability against diverse perturbations and clean inputs (Figure~\ref{fig:e2e_st}). When we adversarially train the searched architectures from the DARTS search space, CRoZe achieves the highest average robustness against 5 different adversarial attacks, with much lower search costs compared to robust one-shot NAS on CIFAR-10 (Table~\ref{tbl:e2e_adv}, Figure~\ref{fig:e2e_at}).

\vspace{-0.05in}
\subsection{Further Analysis}
\label{sec:analysis}
\vspace{-0.05in}
\paragraph{Ablation of Each Component of CRoZe.}
\begin{table}[t]
\caption{\small Comparisons of the final performance of the searched network in NAS-Bench-201 and DARTS search space on CIFAR-10. All models are standard-trained. \textbf{Bold} and \underline{underline} stands for the best and second.}
    \centering
    \resizebox{0.9\textwidth}{!}{
        \begin{tabular}{cccccccacccca}
            \toprule
            
            \multicolumn{3}{c}{Proxy components} & \multicolumn{5}{c}{NAS-Bench-201} & \multicolumn{5}{c}{DARTS} \\
            \cmidrule(l{2pt}r{2pt}){1-3} \cmidrule(l{2pt}r{2pt}){4-8} \cmidrule(l{2pt}r{2pt}){9-13}
            Feature & Parameter & Gradient & Clean & CC. & FGSM & HRS & Avg. & Clean & CC. & FGSM & HRS & Avg. \\ 
            \midrule
            \midrule
            $\checkmark$ & -- & -- & 93.30 & 55.62 & 44.30 & 60.08 & 63.33 & 94.37 & 72.26 & 16.87 & 28.62 & 53.78 \\
            $\checkmark$ & $\checkmark$ & -- & 93.70 & 56.93 & 45.80 & 61.53 & 64.49 & \textbf{94.99} & 74.06 & 16.82 & 28.58 & 53.86 \\
            $\checkmark$ & -- & $\checkmark$ & 93.70 & 56.93 & 45.80 & 61.53  & 64.49 & 94.30 & \textbf{74.91} & 16.67 & 28.33 & 53.55\\
            -- & $\checkmark$ & $\checkmark$ & 93.70 & 56.93 & 45.80 & 61.53  & 64.49 & 94.34 & 74.46 & 15.71 & 26.93 & 52.61 \\
            $\checkmark$ & $\checkmark$ & $\checkmark$ & 93.70 & 56.93 & 45.80 & 61.53  & 64.49  & \underline{94.45} & \underline{74.63} & \textbf{22.38} & 
            \textbf{36.19} & \textbf{56.91} \\
            \midrule
            \bottomrule
            \end{tabular}}
    \label{tbl:ablation_comp}
    \vspace{-0.1in}
\end{table}
\label{sec:ablation_components}
Our proxy consists of three components: feature, parameter, and gradient consistency. To analyze the importance of considering all factors for evaluating the robustness of the given neural architecture, we conduct an ablation study in both the NAS-Bench-201 search space and the DARTS search space. On the NAS-Bench-201 search space, all ablation proxies identified the same candidate architecture except for the feature-based proxy. To compare the ablation proxies thoroughly, we extend our examination to the DARTS search space, which contains a significantly larger number of candidate architectures (e.g., $10^{19}$) compared to the NAS-Bench-201 search space. As shown in Table~\ref{tbl:ablation_comp}, architecture identified solely by feature consistency exhibits better average performance compared to those discerned by proxies without feature consistency, emphasizing the influential role of feature consistency in evaluating robustness. Notably, the architecture selected by our proxy showcases the highest average performance. This indicates that both parameter and gradient consistency can bolster feature consistency, offering more insight into the learning trajectory, as elaborated in Section~\ref{sec:method_croze}. Overall, considering all components is crucial to evaluate the robustness within a single batch.

\vspace{-0.05in}
\paragraph{Feature Variance of the Robust Models.}
\begin{wrapfigure}[8]{r}{0.45\textwidth}
    \centering
    \vspace{-0.18in}
    \includegraphics[width=\linewidth]{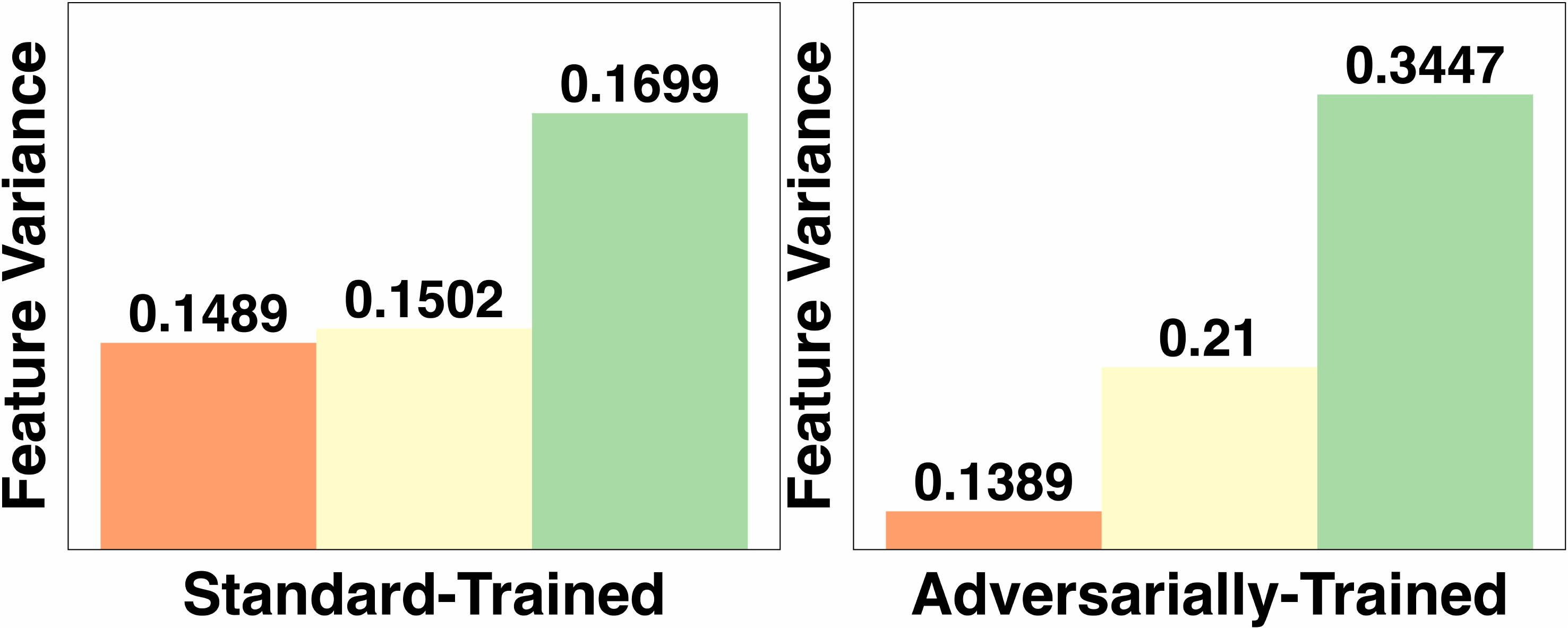}
    \vspace{-0.2in}
    \caption{\small Feature variance across perturbations.}
    \label{fig:feat_dist}
\end{wrapfigure}

Our proxy searches for architectures that can generalize to all types of perturbed inputs, which are not overfitted to a single type of perturbation, as described in Section~\ref{sec:rob_archs} (Eq.~\ref{eq:feature_bound}). To validate this, we analyzed the feature variance of neural architectures that demonstrated the highest performance against a single type of perturbation, namely FGSM and PGD, compared to an architecture selected with the CRoZe proxy from a pool of 300 standard-trained and adversarially-trained models (Figure~\ref{fig:feat_dist}). Features were obtained from 18 different perturbations, including clean, PGD, FGSM, and 15 types of common corruptions. Remarkably, the neural architectures selected with our proxy exhibited the smallest standard deviations between the features obtained from the 18 different perturbations, in both standard and adversarially-trained scenarios. Conversely, the architecture with the best PGD robustness demonstrated 2.48 times larger variations, indicating a higher risk of overfitting to specific perturbations. These results provide compelling evidence of the effectiveness of CRoZe in identifying robust architectures that can consistently extract features and generalize across diverse perturbations.

\vspace{-0.05in}
\paragraph{Assessment of CRoZe Predictiveness.}
\label{sec:predictiveness}
\begin{figure}
    \centering
    \begin{subfigure}[b]{0.24\linewidth}
     \centering
         \includegraphics[width=\linewidth]{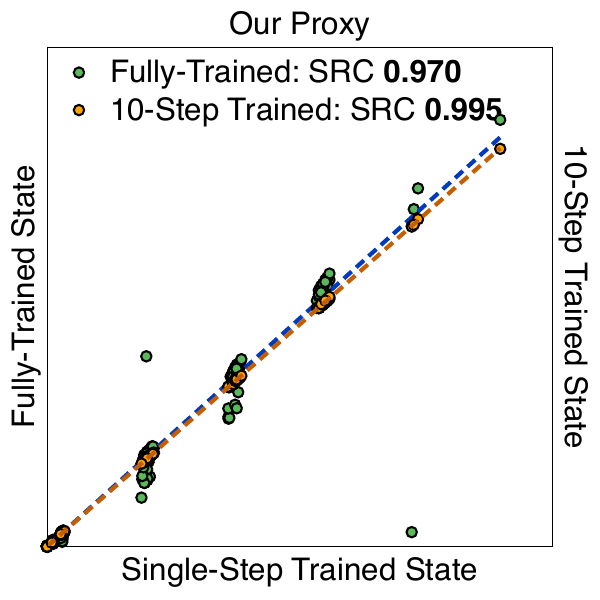}
         \vspace{-0.15in}
         \vspace{-0.1in}
         \label{fig:proxysim}
     \end{subfigure}
     \begin{subfigure}[b]{0.24\linewidth}
         \centering
         \includegraphics[width=\linewidth]{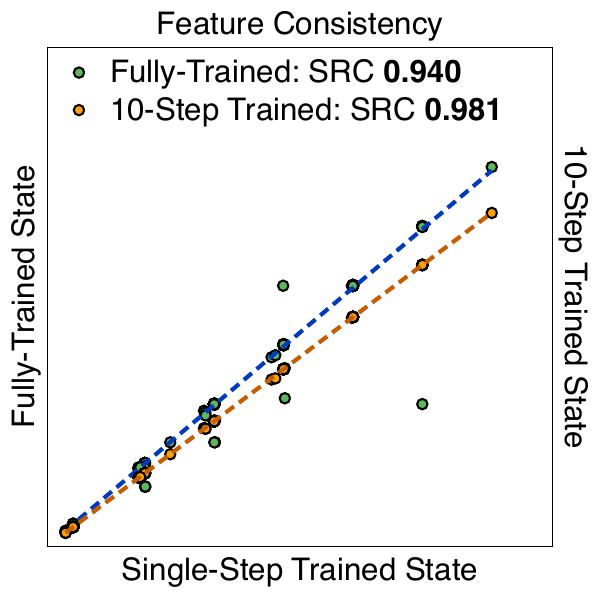}
         \vspace{-0.15in}
         \vspace{-0.1in}
         \label{fig:featsim}
     \end{subfigure}
     \begin{subfigure}[b]{0.24\linewidth}
         \centering
         \includegraphics[width=\linewidth]{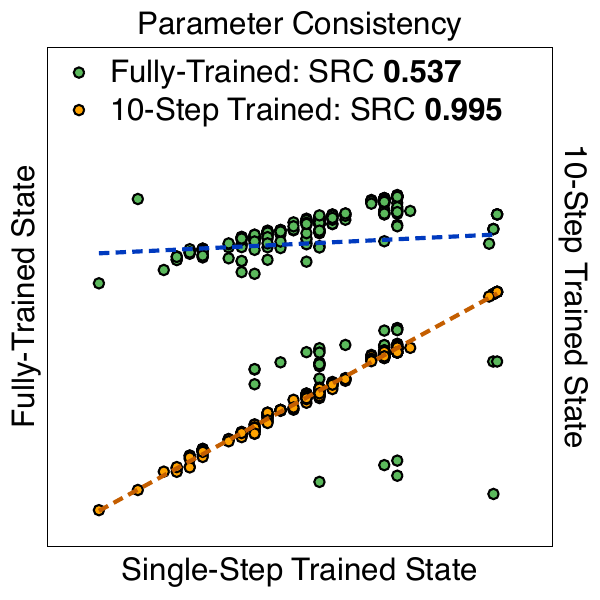}
         \vspace{-0.15in}
         \vspace{-0.1in}
         \label{fig:parmsim}
     \end{subfigure}
     \begin{subfigure}[b]{0.24\linewidth}
         \centering
         \includegraphics[width=\linewidth]{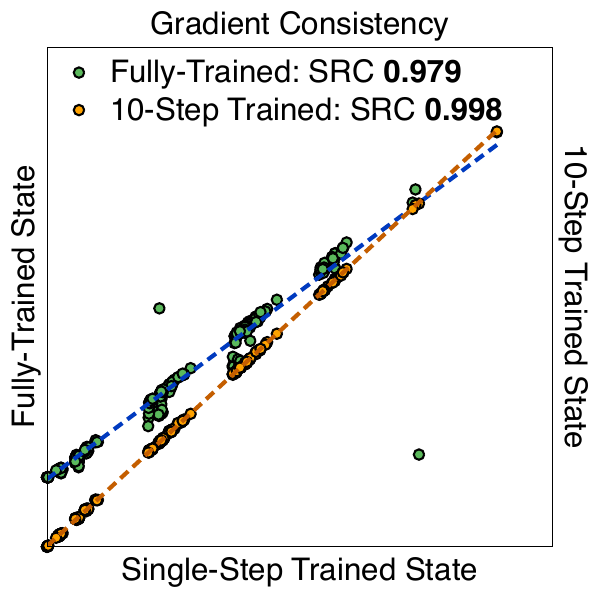}
         \vspace{-0.15in}
         \vspace{-0.1in}
         \label{fig:gradsim}
     \end{subfigure}
    \caption{\small Spearman's $\rho$ in our proxy and each consistency component between the architectures with single-step trained states and two of fully-trained and 10-step trained states, respectively, against clean and perturbed images.}
    \label{fig:proxy_consistency}
    \vspace{-0.2in}
\end{figure}

To empirically validate whether our proxy obtained from a random state accurately represents the proxy in the fully-trained model, we conducted an analysis using Spearman's $\rho$. We randomly sampled 300 architectures from the NAS-Bench-201 search space and trained them on the entire dataset, using both the full training and reduced training only of 10 steps. The Spearman's $\rho$ between the proxy value obtained from a single-step trained state and the 10-step trained state shows a strong correlation of $0.995$. Even after full training, the correlation remained high at $0.970$ (Figure~\ref{fig:proxy_consistency}). These suggest that our proxy, derived from estimated surrogate networks, can significantly reduce the computational costs associated with obtaining fully-trained models. Furthermore, each component of CRoZe consistently demonstrates a high correlation with the fully-trained states, supporting the notion that the high correlation computed with the final state is not solely a result of the collective influence of components, but rather an accurate estimation.

\vspace{-0.05in}
\paragraph{Validation of CRoZe for Estimating Fully-Trained Neural Architectures.}
\label{sec:observation}
\begin{wrapfigure}[8]{r}{0.5\textwidth}
    \centering
    \vspace{-0.25in}
    \includegraphics[width=\linewidth]{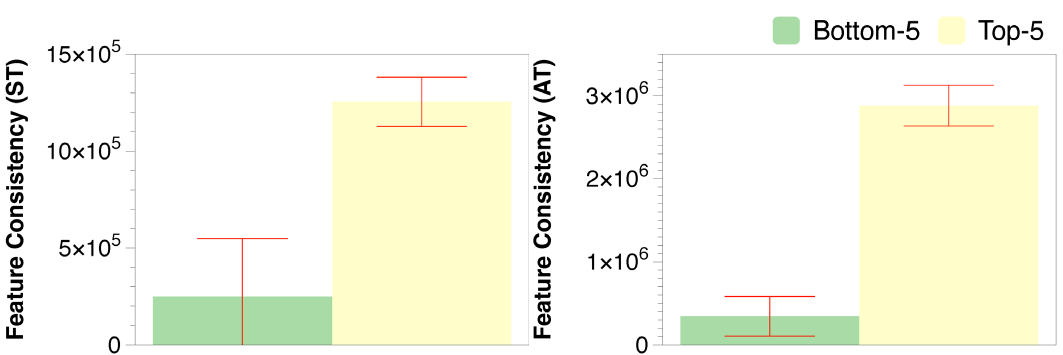}
    \caption{\small Comparisons of feature consistency in fully-trained states.}
    \label{fig:observation}
\end{wrapfigure}

To investigate whether the architectures selected with our proxy possess the desired robustness properties discussed in Section~\ref{sec:rob_archs}, we conduct an analysis. From a pool of 300 samples in the NAS-Bench-201 search space, we select top-5 and bottom-5 architectures based on our proxy, excluding those with a robust accuracy of less than 20\%. We evaluate the performances of these architectures that are standard-trained (ST) and adversarially-trained (AT) against clean inputs and three distinct types of perturbed inputs: adversarial perturbations (i.e., FGSM and PGD), and common corruptions.

\begin{wraptable}{r}{0.5\textwidth}
    \centering
    \vspace{-0.13in}
    \resizebox{\linewidth}{!}{
    \begin{tabular}{lccccc}
    \toprule
    & \multicolumn{5}{c}{Standard-Trained (ST)} \\
    \cmidrule{2-6}
    Type & Clean & CC. & PGD & FGSM & Avg. Rob. \\
    \midrule
    \midrule
    Top-5 & 89.62 & 77.68 & 26.09 & 61.08 & 54.95 \\
    Bottom-5 & 61.34 & 49.13 & 19.35 & 17.01 & 28.50 \\
    \midrule
     & \multicolumn{5}{c}{Adversarially-Trained (AT)} \\
     \cmidrule{2-6}
     & Clean & PGD & HRS(PGD) & FGSM & HRS(FGSM) \\
    \midrule
    Top-5 & 84.36 & 40.55 & 54.76 & 66.48 & 74.34 \\
    Bottom-5 & 51.64 & 22.59 & 31.36 & 37.20 & 43.04 \\
    \midrule
    \bottomrule
    \end{tabular}}
    \caption{\small Comparisons of the performance.}
    \label{tbl:observation_e2e}
    \vspace{-0.2in}
\end{wraptable}

Notably, the top-5 networks show impressive average clean accuracies of 89.62\%, and 84.36\%  on CIFAR-10, while the bottom-5 networks achieve significantly lower average clean accuracies of 61.34\%, 51.64\% in standard training and adversarial training, respectively (Table~\ref{tbl:observation_e2e}). Moreover, the top-5 networks exhibit an average robust accuracy of 54.95\% against perturbations, compared to only 28.50\% for the bottom-5 group. Furthermore, by considering feature consistency between clean and perturbed inputs, as defined in Eq.~\ref{eq:feature_bound}, the top-5 group exhibited similar features, whereas the bottom-5 group showed dissimilar features (Figure~\ref{fig:observation}). This supports that architectures with consistent features across inputs are more likely to make accurate predictions under perturbations. Overall, our analysis provides strong evidence that our proxy can search for generalizable robust architectures that accurately predict the final accuracies under diverse perturbations.

\vspace{-0.05in}
\section{Conclusion}
\label{sec:conclusion}
While neural architecture search (NAS) is a powerful technique for automatically discovering high-performing deep learning models, previous NAS works suffer from two major drawbacks: computational inefficiency and compromised robustness against diverse perturbations, which hinder their applications in real-world scenarios with safety-critical applications.
In this paper, we proposed a simple yet effective lightweight robust NAS framework that can rapidly search for well-generalized neural architectures against diverse tasks and perturbations. To this end, we proposed a novel consistency-based zero-cost proxy that evaluates the robustness of randomly initialized neural networks by measuring the consistency in their features, parameters, and gradients for both clean and perturbed inputs. Experimental results demonstrate the effectiveness of our approach in discovering well-generalized architectures across diverse search spaces, multiple datasets, and various types of perturbations, outperforming the baselines with significantly reduced search costs. Such simplicity and effectiveness of our approach open up new possibilities for automatically discovering high-performing models that are well-suited for safety-critical applications.
\section*{Acknowledgement}
This work was supported by the Institute of Information \& communications Technology Planning \& Evaluation (IITP) grant funded by the Korea government (MSIT) (No.2020-0-00153) and by the Institute of Information \& communications Technology Planning \& Evaluation (IITP) grant funded by the Korea government(MSIT) (No.2019-0-00075, Artificial Intelligence Graduate School Program(KAIST)). We thank Jin Myung Kwak, Eunji Ko, Yulmu Kim, Sohyun An, and Hayeon Lee for providing helpful feedback and suggestions in preparing an earlier version of the manuscript. We also thank the anonymous reviewers for their insightful comments and suggestions.



\bibliography{references}

\begin{thebibliography}{46}
\providecommand{\natexlab}[1]{#1}
\providecommand{\url}[1]{\texttt{#1}}
\expandafter\ifx\csname urlstyle\endcsname\relax
  \providecommand{\doi}[1]{doi: #1}\else
  \providecommand{\doi}{doi: \begingroup \urlstyle{rm}\Url}\fi

\bibitem[Abdelfattah et~al.(2021)Abdelfattah, Mehrotra, Dudziak, and Lane]{abdelfattah2021zero}
Mohamed~S Abdelfattah, Abhinav Mehrotra, {\L}ukasz Dudziak, and Nicholas~D Lane.
\newblock Zero-cost proxies for lightweight nas.
\newblock \emph{International Conference on Learning Representations}, 2021.

\bibitem[Baker et~al.(2017)Baker, Gupta, Naik, and Raskar]{baker2016}
Bowen Baker, Otkrist Gupta, Nikhil Naik, and Ramesh Raskar.
\newblock Designing neural network architectures using reinforcement learning.
\newblock In \emph{International Conference on Learning Representations}, 2017.

\bibitem[Biggio et~al.(2013)Biggio, Corona, Maiorca, Nelson, {\v{S}}rndi{\'c}, Laskov, Giacinto, and Roli]{biggio2013evasion}
Battista Biggio, Igino Corona, Davide Maiorca, Blaine Nelson, Nedim {\v{S}}rndi{\'c}, Pavel Laskov, Giorgio Giacinto, and Fabio Roli.
\newblock Evasion attacks against machine learning at test time.
\newblock In \emph{Joint European conference on machine learning and knowledge discovery in databases}, pages 387--402. Springer, 2013.

\bibitem[Cai et~al.(2019)Cai, Gan, Wang, Zhang, and Han]{cai2019once}
Han Cai, Chuang Gan, Tianzhe Wang, Zhekai Zhang, and Song Han.
\newblock Once-for-all: Train one network and specialize it for efficient deployment.
\newblock \emph{International Conference on Learning Representations}, 2019.

\bibitem[Carlini and Wagner(2017)]{carlini2017cw}
Nicholas Carlini and David Wagner.
\newblock Towards evaluating the robustness of neural networks.
\newblock In \emph{IEEE symposium on security and privacy (sp)}, pages 39--57. IEEE, 2017.

\bibitem[Chen et~al.(2020)Chen, Wang, Cheng, Tang, and Hsieh]{chen2020drnas}
Xiangning Chen, Ruochen Wang, Minhao Cheng, Xiaocheng Tang, and Cho-Jui Hsieh.
\newblock Drnas: Dirichlet neural architecture search.
\newblock \emph{arXiv preprint arXiv:2006.10355}, 2020.

\bibitem[Croce and Hein(2020)]{croce2020reliable}
Francesco Croce and Matthias Hein.
\newblock Reliable evaluation of adversarial robustness with an ensemble of diverse parameter-free attacks.
\newblock In \emph{ICML}, pages 2206--2216. PMLR, 2020.

\bibitem[Cubuk et~al.(2020)Cubuk, Zoph, Shlens, and Le]{cubuk2020randaugment}
Ekin~D Cubuk, Barret Zoph, Jonathon Shlens, and Quoc~V Le.
\newblock Randaugment: Practical automated data augmentation with a reduced search space.
\newblock In \emph{Advances in Neural Information Processing Systems}, 2020.

\bibitem[Devaguptapu et~al.(2021)Devaguptapu, Agarwal, Mittal, Gopalani, and Balasubramanian]{devaguptapu2020empirical}
Chaitanya Devaguptapu, Devansh Agarwal, Gaurav Mittal, Pulkit Gopalani, and Vineeth~N Balasubramanian.
\newblock On adversarial robustness: A neural architecture search perspective.
\newblock In \emph{IEEE International Conference on Computer Vision}, pages 152--161, 2021.

\bibitem[Dodge and Karam(2017)]{dodge2017study}
Samuel Dodge and Lina Karam.
\newblock A study and comparison of human and deep learning recognition performance under visual distortions.
\newblock In \emph{2017 26th international conference on computer communication and networks (ICCCN)}, pages 1--7. IEEE, 2017.

\bibitem[Dong et~al.(2023)Dong, Li, and Wei]{dong2023diswot}
Peijie Dong, Lujun Li, and Zimian Wei.
\newblock Diswot: Student architecture search for distillation without training.
\newblock \emph{IEEE Conference on Computer Vision and Pattern Recognition}, 2023.

\bibitem[Dong and Yang(2019)]{nasbench201}
Xuanyi Dong and Yi~Yang.
\newblock Searching for a robust neural architecture in four gpu hours.
\newblock In \emph{IEEE Conference on Computer Vision and Pattern Recognition}, pages 1761--1770, 2019.

\bibitem[Elsken et~al.(2019)Elsken, Metzen, and Hutter]{elsken2018efficient}
Thomas Elsken, Jan~Hendrik Metzen, and Frank Hutter.
\newblock Efficient multi-objective neural architecture search via lamarckian evolution.
\newblock \emph{International Conference on Learning Representations}, 2019.

\bibitem[Foret et~al.(2021)Foret, Kleiner, Mobahi, and Neyshabur]{foret2021sharpnessaware}
Pierre Foret, Ariel Kleiner, Hossein Mobahi, and Behnam Neyshabur.
\newblock Sharpness-aware minimization for efficiently improving generalization.
\newblock In \emph{International Conference on Learning Representations}, 2021.

\bibitem[Glorot and Bengio(2010)]{glorot2010understanding}
Xavier Glorot and Yoshua Bengio.
\newblock Understanding the difficulty of training deep feedforward neural networks.
\newblock In \emph{Proceedings of the thirteenth international conference on artificial intelligence and statistics}, pages 249--256. JMLR Workshop and Conference Proceedings, 2010.

\bibitem[Goodfellow et~al.(2015)Goodfellow, Shlens, and Szegedy]{goodfellow2014fgsm}
Ian~J Goodfellow, Jonathon Shlens, and Christian Szegedy.
\newblock Explaining and harnessing adversarial examples.
\newblock In \emph{International Conference on Learning Representations}, 2015.

\bibitem[Gubri et~al.(2022)Gubri, Cordy, Papadakis, Traon, and Sen]{gubri2022lgv}
Martin Gubri, Maxime Cordy, Mike Papadakis, Yves~Le Traon, and Koushik Sen.
\newblock Lgv: Boosting adversarial example transferability from large geometric vicinity.
\newblock In \emph{ECCV}, pages 603--618. Springer, 2022.

\bibitem[Guo et~al.(2020)Guo, Yang, Xu, Liu, and Lin]{guo2019meets}
Minghao Guo, Yuzhe Yang, Rui Xu, Ziwei Liu, and Dahua Lin.
\newblock When nas meets robustness: In search of robust architectures against adversarial attacks.
\newblock \emph{IEEE Conference on Computer Vision and Pattern Recognition}, 2020.

\bibitem[He et~al.(2015)He, Zhang, Ren, and Sun]{he2015delving}
Kaiming He, Xiangyu Zhang, Shaoqing Ren, and Jian Sun.
\newblock Delving deep into rectifiers: Surpassing human-level performance on imagenet classification.
\newblock In \emph{Proceedings of the IEEE international conference on computer vision}, pages 1026--1034, 2015.

\bibitem[Hendrycks and Dietterich(2019)]{hendrycks2019benchmarking}
Dan Hendrycks and Thomas Dietterich.
\newblock Benchmarking neural network robustness to common corruptions and perturbations.
\newblock \emph{International Conference on Learning Representations}, 2019.

\bibitem[Hendrycks et~al.(2020)Hendrycks, Mu, Cubuk, Zoph, Gilmer, and Lakshminarayanan]{hendrycks2019augmix}
Dan Hendrycks, Norman Mu, Ekin~D Cubuk, Barret Zoph, Justin Gilmer, and Balaji Lakshminarayanan.
\newblock Augmix: A simple data processing method to improve robustness and uncertainty.
\newblock \emph{International Conference on Learning Representations}, 2020.

\bibitem[Jung et~al.(2023)Jung, Lukasik, and Keuper]{jung2023neural}
Steffen Jung, Jovita Lukasik, and Margret Keuper.
\newblock Neural architecture design and robustness: A dataset.
\newblock In \emph{International Conference on Learning Representations}, 2023.

\bibitem[Lee et~al.(2019)Lee, Ajanthan, and Torr]{lee2018snip}
Namhoon Lee, Thalaiyasingam Ajanthan, and Philip~HS Torr.
\newblock Snip: Single-shot network pruning based on connection sensitivity.
\newblock \emph{International Conference on Learning Representations}, 2019.

\bibitem[Liu et~al.(2018{\natexlab{a}})Liu, Zoph, Neumann, Shlens, Hua, Li, Fei-Fei, Yuille, Huang, and Murphy]{liu2018progressive}
Chenxi Liu, Barret Zoph, Maxim Neumann, Jonathon Shlens, Wei Hua, Li-Jia Li, Li~Fei-Fei, Alan Yuille, Jonathan Huang, and Kevin Murphy.
\newblock Progressive neural architecture search.
\newblock In \emph{European Conference on Computer Vision}, pages 19--34, 2018{\natexlab{a}}.

\bibitem[Liu et~al.(2018{\natexlab{b}})Liu, Simonyan, Vinyals, Fernando, and Kavukcuoglu]{liu2017hierarchical}
Hanxiao Liu, Karen Simonyan, Oriol Vinyals, Chrisantha Fernando, and Koray Kavukcuoglu.
\newblock Hierarchical representations for efficient architecture search.
\newblock In \emph{International Conference on Learning Representations}, 2018{\natexlab{b}}.

\bibitem[Liu et~al.(2019)Liu, Simonyan, and Yang]{liu2018darts}
Hanxiao Liu, Karen Simonyan, and Yiming Yang.
\newblock Darts: Differentiable architecture search.
\newblock \emph{International Conference on Learning Representations}, 2019.

\bibitem[Liu et~al.(2021)Liu, Zhang, Kuang, Zhou, Xue, Wang, Chen, Yang, Liao, and Zhang]{liu2021fisher}
Liyang Liu, Shilong Zhang, Zhanghui Kuang, Aojun Zhou, Jing-Hao Xue, Xinjiang Wang, Yimin Chen, Wenming Yang, Qingmin Liao, and Wayne Zhang.
\newblock Group fisher pruning for practical network compression.
\newblock In \emph{ICML}, pages 7021--7032. PMLR, 2021.

\bibitem[Luo et~al.(2018)Luo, Tian, Qin, Chen, and Liu]{luo2018neural}
Renqian Luo, Fei Tian, Tao Qin, Enhong Chen, and Tie-Yan Liu.
\newblock Neural architecture optimization.
\newblock \emph{Advances in Neural Information Processing Systems}, 31, 2018.

\bibitem[Madry et~al.(2018)Madry, Makelov, Schmidt, Tsipras, and Vladu]{madry2017pgd}
Aleksander Madry, Aleksandar Makelov, Ludwig Schmidt, Dimitris Tsipras, and Adrian Vladu.
\newblock Towards deep learning models resistant to adversarial attacks.
\newblock In \emph{International Conference on Learning Representations}, 2018.

\bibitem[Mellor et~al.(2021)Mellor, Turner, Storkey, and Crowley]{mellor2021neural}
Joe Mellor, Jack Turner, Amos Storkey, and Elliot~J Crowley.
\newblock Neural architecture search without training.
\newblock In \emph{International Conference on Machine Learning}, pages 7588--7598. PMLR, 2021.

\bibitem[Mok et~al.(2021)Mok, Na, Choe, and Yoon]{mok2021advrush}
Jisoo Mok, Byunggook Na, Hyeokjun Choe, and Sungroh Yoon.
\newblock Advrush: Searching for adversarially robust neural architectures.
\newblock In \emph{IEEE Conference on Computer Vision and Pattern Recognition}, pages 12322--12332, 2021.

\bibitem[Moosavi-Dezfooli et~al.(2016)Moosavi-Dezfooli, Fawzi, and Frossard]{moosavi2016deepfool}
Seyed-Mohsen Moosavi-Dezfooli, Alhussein Fawzi, and Pascal Frossard.
\newblock Deepfool: a simple and accurate method to fool deep neural networks.
\newblock In \emph{Proceedings of the IEEE conference on computer vision and pattern recognition}, pages 2574--2582, 2016.

\bibitem[Pham et~al.(2018)Pham, Guan, Zoph, Le, and Dean]{pham2018efficient}
Hieu Pham, Melody Guan, Barret Zoph, Quoc Le, and Jeff Dean.
\newblock Efficient neural architecture search via parameters sharing.
\newblock In \emph{International Conference on Machine Learning}, pages 4095--4104. PMLR, 2018.

\bibitem[Real et~al.(2017)Real, Moore, Selle, Saxena, Suematsu, Tan, Le, and Kurakin]{real2017large}
Esteban Real, Sherry Moore, Andrew Selle, Saurabh Saxena, Yutaka~Leon Suematsu, Jie Tan, Quoc~V Le, and Alexey Kurakin.
\newblock Large-scale evolution of image classifiers.
\newblock In \emph{ICML}, 2017.

\bibitem[Real et~al.(2019)Real, Aggarwal, Huang, and Le]{real2019regularized}
Esteban Real, Alok Aggarwal, Yanping Huang, and Quoc~V Le.
\newblock Regularized evolution for image classifier architecture search.
\newblock In \emph{AAAI Conference on Artificial Intelligence}, 2019.

\bibitem[Szegedy et~al.(2014)Szegedy, Zaremba, Sutskever, Bruna, Erhan, Goodfellow, and Fergus]{szegedy2013intriguing}
Christian Szegedy, Wojciech Zaremba, Ilya Sutskever, Joan Bruna, Dumitru Erhan, Ian Goodfellow, and Rob Fergus.
\newblock Intriguing properties of neural networks.
\newblock \emph{International Conference on Learning Representations}, 2014.

\bibitem[Tanaka et~al.(2020)Tanaka, Kunin, Yamins, and Ganguli]{tanaka2020synflow}
Hidenori Tanaka, Daniel Kunin, Daniel~L Yamins, and Surya Ganguli.
\newblock Pruning neural networks without any data by iteratively conserving synaptic flow.
\newblock \emph{Advances in Neural Information Processing Systems}, 33:\penalty0 6377--6389, 2020.

\bibitem[Uesato et~al.(2018)Uesato, O’donoghue, Kohli, and Oord]{uesato2018adversarial}
Jonathan Uesato, Brendan O’donoghue, Pushmeet Kohli, and Aaron Oord.
\newblock Adversarial risk and the dangers of evaluating against weak attacks.
\newblock In \emph{ICML}, pages 5025--5034. PMLR, 2018.

\bibitem[Wang et~al.(2020)Wang, Zhang, and Grosse]{wang2020grasp}
Chaoqi Wang, Guodong Zhang, and Roger Grosse.
\newblock Picking winning tickets before training by preserving gradient flow.
\newblock \emph{International Conference on Learning Representations}, 2020.

\bibitem[Wong et~al.(2020)Wong, Rice, and Kolter]{wong2020fast}
Eric Wong, Leslie Rice, and J~Zico Kolter.
\newblock Fast is better than free: Revisiting adversarial training.
\newblock \emph{arXiv preprint arXiv:2001.03994}, 2020.

\bibitem[Wu et~al.(2020)Wu, Xia, and Wang]{awp20wu}
Dongxian Wu, Shu-Tao Xia, and Yisen Wang.
\newblock Adversarial weight perturbation helps robust generalization.
\newblock In H.~Larochelle, M.~Ranzato, R.~Hadsell, M.F. Balcan, and H.~Lin, editors, \emph{Advances in Neural Information Processing Systems}, 2020.

\bibitem[Xu et~al.(2020)Xu, Xie, Zhang, Chen, Qi, Tian, and Xiong]{xu2019pc}
Yuhui Xu, Lingxi Xie, Xiaopeng Zhang, Xin Chen, Guo-Jun Qi, Qi~Tian, and Hongkai Xiong.
\newblock Pc-darts: Partial channel connections for memory-efficient architecture search.
\newblock \emph{International Conference on Learning Representations}, 2020.

\bibitem[Zhang et~al.(2019)Zhang, Yu, Jiao, Xing, Ghaoui, and Jordan]{zhang2019trades}
Hongyang Zhang, Yaodong Yu, Jiantao Jiao, Eric~P Xing, Laurent~El Ghaoui, and Michael~I Jordan.
\newblock Theoretically principled trade-off between robustness and accuracy.
\newblock In \emph{International Conference on Machine Learning}, 2019.

\bibitem[Zhong et~al.(2018)Zhong, Yan, Wu, Shao, and Liu]{zhong2018practical}
Zhao Zhong, Junjie Yan, Wei Wu, Jing Shao, and Cheng-Lin Liu.
\newblock Practical block-wise neural network architecture generation.
\newblock In \emph{IEEE Conference on Computer Vision and Pattern Recognition}, pages 2423--2432, 2018.

\bibitem[Zoph and Le(2017)]{zoph2016neural}
Barret Zoph and Quoc~V Le.
\newblock Neural architecture search with reinforcement learning.
\newblock \emph{International Conference on Learning Representations}, 2017.

\bibitem[Zoph et~al.(2018)Zoph, Vasudevan, Shlens, and Le]{Zoph_2018_CVPR}
Barret Zoph, Vijay Vasudevan, Jonathon Shlens, and Quoc~V. Le.
\newblock Learning transferable architectures for scalable image recognition.
\newblock In \emph{IEEE Conference on Computer Vision and Pattern Recognition}, 2018.

\end{thebibliography}
\newpage
\appendix
\onecolumn
\begin{center}{\bf {\LARGE Supplementary Materials}}\end{center}
\begin{center}{\bf {\Large Generalizable Lightweight Proxy for Robust NAS \\ against Diverse Perturbations}}
\end{center}

\section{Experimental Setting}
\label{sup:exp_setting}

\paragraph{Search space}
\label{sec:supple_searchspace}
Based on the cell-based neural architecture search space~\cite{Zoph_2018_CVPR}, we regard the whole network as the composition of repeated cells. Thus, we search for the optimal cell architectures and stack them repeatedly to construct the entire network. In the cell-based search phase, each cell can be represented as a directed acyclic graph (DAG), which has $N$ nodes that represent the feature maps $z_j (j=1, \cdots, N)$ and each edge between arbitrary node $i$ and node $j$ represents an operation $o_{i,j}$ chosen from the operation pool, where  $o_{i,j} \in ~\mathcal{O} = \{o_k, k=1, 2, \cdots, n\}$. Each feature map $z_j$ is obtained from all of its predecessors as follows:
\begin{equation}
 \label{equation:cell_based_forward}
    x_j = \sum_{i<j} o_{i, j}(x_i)
 \end{equation}
In this work, we utilize NAS-Bench-201~\citep{nasbench201}, and DARTS~\citep{liu2018darts} search space, where different operation pools are used, which are $~\mathcal{O} = \{1 \times 1$ conv., $3 \times 3$ conv., $3\times 3$ avg. pooling, skip, zero$\}$, and $~\mathcal{O} = \{3 \times 3$ conv., $3 \times 3$ dil.conv., $5 \times 5$ conv., $5 \times 5$ dil. conv., $7 \times 7$ conv., $3\times 3$ max pooling, $3\times 3$ avg. pooling, skip, zero$\}$, respectively. Especially, for the NAS-Bench-201 search space, we additionally use the~\citet{jung2023neural} dataset that includes robust accuracies of candidate neural architectures in NAS-Bench-201 search space to demonstrate the efficacy of our proposed proxy regarding searching generalized architectures against diverse perturbations and clean inputs.

\paragraph{Adversarial Evaluation}
To evaluate standard-trained models, we utilize the robust NAS-Bench-201~\citep{jung2023neural} datasets, allowing us to achieve robust accuracy. On the other hand, to evaluate adversarially-trained models, we construct our own dataset, as described in Section~\ref{sec:nasbench201_adv}, enabling us to obtain robust accuracy against adversarial attacks. In all our experiments, we obtain robust accuracy on CIFAR-10 against FGSM attack with an attack size ($\epsilon$) of 8.0/255.0 and attack step size ($\alpha$) of 8.0/2550.0 while we utilize robust accuracies against FGSM attack with attack size ($\epsilon$) of 4.0/255.0 on ImageNet16-120. 

\paragraph{End-to-end sampling}
We sample 5,000 number of neural architectures on the DARTS search space to obtain end-to-end performance on CIFAR-10 and CIFAR-100. For both CIFAR-10 and CIFAR-100, we utilize the AE+warmup+move sampling strategy described in Section~\ref{sec:e2e_darts}, where our proxy guides the sampling towards the pool of architectures with high proxy values. Specifically, for CIFAR-10, we use an init pool (e.g., warmup) of 3,000 and a sample pool for AE (e.g., move) of 50. On the other hand, we employ an init pool of 3,000 and a sample pool of 100 for CIFAR-100.

\section{Experimental Results}
\subsection{Ablation on Each Component of CRoZe}
\label{sec:sup_ablation_proxy}

In Section~\ref{sec:method_croze}, we introduce our proxy, which consists of three components: feature, parameter, and gradient consistency. We discuss the importance of considering all three components to accurately evaluate the robustness of neural architectures in a random state (Section~\ref{sec:ablation_components}). To further analyze the contributions of each factor, we conduct an ablation study in both the NAS-Bench-201 search space and the DARTS search space.

We find that relying solely on feature consistency in a random state is insufficient to evaluate the robustness of architectures. The proxy with only feature consistency shows a lower correlation in both standard training and adversarial training scenarios compared to CRoZe in the NAS-Bench-201 search space (Table~\ref{tbl:sup_src_nasbench201} and Table~\ref{tbl:sup_src_nasbench201_fastat}). This indicates that high scores obtained based on feature consistency on a single batch may not accurately reflect the performance across the entire dataset. On the other hand, when parameter or gradient similarity is added to the proxy, the correlation consistently improves, suggesting that these factors complement feature consistency by imposing stricter constraints on the parameter space and convergence directions, respectively. While the proxy considering only parameter and gradient similarity achieves better Spearman's $\rho$ compared to our proxy, the top-3 architectures chosen by our proxy exhibit higher average performance than those discovered by the proxy without feature consistency (Table~\ref{tbl:sup_e2e_nasbench}). 

\hbadness=99999  
\begin{table}[t]
\caption{\small Comparison of Spearman's $\rho$ between the actual accuracies and the proxy values on CIFAR-10 in the NAS-Bench-201 search space. Clean stands for clean accuracy and robust accuracies are evaluated against adversarial perturbations (FGSM~\cite{goodfellow2014fgsm}) and common corruptions (CC.~\citep{hendrycks2019benchmarking}). Avg. stands for average Spearman's $\rho$ values with all accuracies.}
    \begin{center}
    \begin{subtable}[t]{0.43\textwidth}
    \vspace{-0.05in}
    \resizebox{\textwidth}{!}{
        \begin{tabular}{cccccca}
            \toprule
            \multicolumn{3}{c}{Proxy components} & \multicolumn{4}{c}{Standard-Trained} \\
            \cmidrule(l{2pt}r{2pt}){1-3} \cmidrule(l{2pt}r{2pt}){4-7}
            Feature & Parameter & Gradient & Clean & FGSM & CC. & Avg. \\ 
            \midrule
            \midrule
            $\checkmark$ & -- & -- & 0.718 & 0.701 & 0.341 & 0.587 \\
            $\checkmark$ & $\checkmark$ & -- & 0.750 & 0.762 & 0.384 & 0.632 \\
            $\checkmark$ & -- & $\checkmark$ & 0.822 & 0.824 & 0.434 & 0.693 \\
            -- & $\checkmark$ & $\checkmark$ & 0.824 & 0.827 & 0.437 & 0.696 \\
            $\checkmark$ & $\checkmark$ & $\checkmark$ & 0.823 & 0.826 & 0.436 & 0.695 \\
            \midrule
            \bottomrule
            \end{tabular}
            }
            \vspace{-0.1in}
            \caption{Standard-Trained}
            \label{tbl:sup_src_nasbench201}
    \end{subtable}
    \begin{subtable}[t]{0.56\textwidth}
    \vspace{-0.05in}
        \resizebox{\textwidth}{!}{
            \begin{tabular}{cccccccc}
                \toprule
                \multicolumn{3}{c}{Proxy components} & \multicolumn{5}{c}{Adversarially-Trained} \\
                \cmidrule(l{2pt}r{2pt}){1-3} \cmidrule(l{2pt}r{2pt}){4-8} 
                Feature & Parameter & Gradient & Clean & FGSM & PGD & HRS(FGSM) & HRS(PGD) \\
                \midrule
                \midrule
                $\checkmark$ & -- & -- & 0.602 & 0.295 & 0.329 & 0.442 & 0.404 \\
                $\checkmark$ & $\checkmark$ & -- & 0.677 & 0.343 & 0.431 & 0.542 & 0.527 \\
                $\checkmark$ & -- & $\checkmark$ & 0.707 & 0.405 & 0.489 & 0.587 & 0.573 \\
                -- & $\checkmark$ & $\checkmark$ & 0.731 & 0.422 & 0.507 & 0.610 & 0.595 \\
                $\checkmark$ & $\checkmark$ & $\checkmark$ & 0.723 & 0.417 & 0.501 & 0.602 & 0.588 \\
                \midrule
                \bottomrule
                \end{tabular} 
        }
        \vspace{-0.1in}
        \caption{Adversarially-Trained}
        \label{tbl:sup_src_nasbench201_fastat}
    \end{subtable}
    \end{center}
    \vspace{-0.3in}
\end{table}

We further conduct ablation experiments on CIFAR-10 in the DARTS search space, which contains about $10^{19}$ number of candidate architectures that is significantly larger than the NAS-Bench-201 search space containing $15625$ architectures. The proxy without feature consistency yielded architectures with poor robust accuracies, while the architectures selected by our proxy consistently outperformed the former on both clean and perturbed images (Table~\ref{tbl:sup_e2e_darts}). Furthermore, architecture identified solely by feature consistency exhibits better average performance compared to those discerned by proxies without feature consistency. This clearly demonstrates the influential role of feature consistency in evaluating robustness. Overall, our proposed proxy effectively searches high-performing architectures by employing consistency across features, parameters, and gradients to estimate the robustness of the given architectures within a single gradient step. The overall algorithm of CRoZe is described in Algorithm~\ref{algo:algorithm}.

\begin{table}[t]
\caption{\small Comparisons of the final performance of the searched network in NAS-Bench-201 and DARTS search space on CIFAR-10. \textbf{Bold} and \underline{underline} stands for the best and second.}
    \begin{center}
    \begin{subtable}[t]{0.56\textwidth}
    \resizebox{\textwidth}{!}{
        \begin{tabular}{ccccccaccca}
            \toprule
            \multicolumn{3}{c}{Proxy components} & \multicolumn{4}{c}{Standard-Trained (Top-1)} & \multicolumn{4}{c}{Standard-Trained (Top-3)}\\
            \cmidrule(l{2pt}r{2pt}){1-3} \cmidrule(l{2pt}r{2pt}){4-7} \cmidrule(l{2pt}r{2pt}){8-11}
            Feature & Parameter & Gradient & Clean & FGSM & CC. & Avg. & Clean & FGSM & CC. & Avg.\\ 
            \midrule
            \midrule
            $\checkmark$ & -- & -- & 93.30 & 44.30 & 55.62 & 64.41 & 92.93 & 37.60 & 54.03 & 61.52\\
            $\checkmark$ & $\checkmark$ & -- & 93.70 & 45.80 & 56.93 & 65.48 & 93.63 & 48.20 & 55.39 & \textbf{65.74}\\
            $\checkmark$ & -- & $\checkmark$ & 93.70 & 45.80 & 56.93 & 65.48 & 93.63 & 48.20 & 55.39 & \textbf{65.74}\\
            -- & $\checkmark$ & $\checkmark$ & 93.70 & 45.80 & 56.93 & 65.48 & 93.43 & 41.37 & 55.30 & 63.37\\
            $\checkmark$ & $\checkmark$ & $\checkmark$ & 93.70 & 45.80 & 56.93 & 65.48 & 93.93 & 43.87 & 56.11 & \underline{64.64}\\
            \midrule
            \bottomrule
            \end{tabular}
            }
            \vspace{-0.1in}
            \caption{NAS-Bench-201 search space}
            \label{tbl:sup_e2e_nasbench}
    \end{subtable}
    \begin{subtable}[t]{0.42\textwidth}
        \centering
        \resizebox{\textwidth}{!}{
        \begin{tabular}{ccccccca}
            \toprule
            \multicolumn{3}{c}{Proxy Components} & \multicolumn{5}{c}{Standard-Trained} \\ 
            \cmidrule(r){1-3} \cmidrule(r){4-8} 
            Feature & Parameter & Gradient & Clean & CC. & FGSM & HRS & Avg. \\ 
            \midrule
            \midrule
            $\checkmark$ & -- & -- & 94.37 & 72.26 & \underline{16.87} & \underline{28.62} & 53.78 \\
            $\checkmark$ & $\checkmark$ & -- & \textbf{94.99} & 74.06 & 16.82 & 28.58 & \underline{53.86} \\
            $\checkmark$ & -- & $\checkmark$ & 94.30 & \textbf{74.91} & 16.67 & 28.33 & 53.55 \\
            -- & $\checkmark$ & $\checkmark$ & 94.34 & 74.46 & 15.71 & 26.93 & 52.61 \\
            $\checkmark$ & $\checkmark$ & $\checkmark$ & \underline{94.45} & \underline{74.63} & \textbf{22.38} & \textbf{36.19} & \textbf{56.91} \\
            \midrule
            \bottomrule
        \end{tabular}}
        \vspace{-0.1in}
        \caption{DARTS search space}
        \label{tbl:sup_e2e_darts}
    \end{subtable}
    \end{center}
    \label{tbl:sup_e2e}
    \vspace{-0.2in}
\end{table}

\begin{algorithm}
\DontPrintSemicolon
\caption{\textbf{C}nsistency-based \textbf{Ro}bust \textbf{Ze}ro-cost Proxy (\textbf{CRoZe}).}
\KwIn{A single batch of given dataset $B=\{(x, y)\}$, network $f_{\theta}(\cdot)$ consists of $M$ layer, which is architecture $\mathcal{A}$ with parameterized by $\theta$}
\KwOut{Proxy value, \textit{CRoZe}}

\boxit{figred}{0.5}\tcc{Estimate robust network $f_{\theta^r}$ as done in Eq.~\ref{equation:attack_params}.} \;

\For{$m = 1, \cdots, M $}{
$\theta^r_m \leftarrow \theta_m + \beta*\frac{\nabla_{\theta_m}\mathcal{L}\big(f_{\theta}(x), y\big)}{\lVert 
 \nabla_{\theta_m}\mathcal{L}\big(f_{\theta}(x), y\big)\rVert}* \lVert \theta_m\rVert$
}

\BlankLine
\boxit{figgreen}{0.5}
\tcc{Generate perturbed input $x'$ using $f_{\theta^r}$. as done in Eq.~\ref{equation:attack_input}} \;
$\small\delta = \epsilon \mathtt{sign}\Big(\nabla_{x} \mathcal{L}\big(f_{\theta^r}(x), y\big)\Big)$ \;
$x' = x + \delta$
\BlankLine
\boxit{figblue}{1.2}
\tcc{Calculate consistency in features, parameters, and gradients as done in Section~\ref{sec:method_croze}} \;
\tcc{Calculate gradients of both clean network $f_{\theta}$ and robust network $f_{\theta^r}$}
$g = \nabla_{\theta}\mathcal{L}\big(f_{\theta}(x), y)$\;
$g^r = \nabla_{\theta^r}\mathcal{L}\big(f_{\theta^r}(x'), y)$\;
\BlankLine
\tcc{Single gradient step for clean network $f_{\theta}$ and robust network $f_{\theta^r}$}
$\theta_1 \leftarrow \theta - \gamma g$\;
$\theta_1^r \leftarrow \theta^r - \gamma g^r$ \;
\BlankLine
\For{$m = 1, \cdots, M $}{
    \tcc{Feature consistency}
    $\mathcal{Z}_m(f_\theta(x), f_{\theta^r}(x')) = 1+\frac{z_m \cdot z_m^{r}}{\lVert z_m\rVert \lVert z_m^r\rVert}$ \;
    \BlankLine
    \tcc{Parameter consistency}
    $\mathcal{P}_m(\theta_1, \theta_1^r) = 1 + \frac{\theta_{1,m}\cdot \theta_{1,m}^r}{\lVert \theta_{1,m}\rVert \lVert \theta_{1,m}^r\rVert}$
    \BlankLine
    \tcc{Gradient consistency}
    $\mathcal{G}_m(g, g^r) = \big\lvert\frac{g_m \cdot g_m^{r}}{\lVert g_m\rVert \lVert g_m^r\rVert}\big\rvert$
}
\textit{CRoZe} = $\sum_{m=1}^M \mathcal{Z}_m \times \mathcal{P}_m \times \mathcal{G}_m$ \;
\vspace{0.1in}
\textbf{return} \textit{CRoZe};
\label{algo:algorithm}
\end{algorithm}

\begin{table}[ht]
\vspace{-0.1in}
    \begin{center}
    \caption{\small Comparison of Spearman's $\rho$ between the actual accuracies and the proxy values on CIFAR-10, CIFAR-100 and ImageNet16-120 in NAS-Bench-201 search space. Avg. stands for average Spearman's $\rho$ values with all accuracies within each task and CC. stands for the average of 15 different types of common corruption. All models are standard-trained.}
    \label{tbl:sup_input_perturbation_others}
    \resizebox{0.9\linewidth}{!}{
        \small {
        \begin{tabular}{lcccccacccacca}
            \toprule
            \centering
              & \multicolumn{6}{c}{CIFAR-10} & \multicolumn{4}{c}{CIFAR-100} & \multicolumn{3}{c}{ImageNet16-120} \\
            \cmidrule(r){2-7}\cmidrule(r){8-11}\cmidrule(r){12-14}
             &  & \multicolumn{3}{c}{FGSM}&  & \cellcolor{white} &  & FGSM & &  \cellcolor{white}  &  & FGSM & \cellcolor{white} \\ 
             \cmidrule{3-5}\cmidrule{9-9}\cmidrule{13-13}
             Perturbation Type& Clean & $\epsilon=8$ & $\epsilon=4$ & $\epsilon=2$ & CC. & Avg. & Clean & $\epsilon=4$ & CC. & Avg. & Clean & $\epsilon=4$  & Avg. \\
            \midrule
            \midrule
            Gaussian Noise & 0.810 & 0.821 & 0.797 & 0.778 & 0.436 & 0.728 & 0.774 & 0.693 & 0.542 & 0.670 & 0.741 & 0.671 & 0.706 \\
            Adversarial & 0.823 & 0.823 & 0.826 & 0.801 & 0.436 & 0.682 & 0.787 & 0.693 & 0.533 & 0.671 & 0.769 & 0.696 & 0.733 \\
            \midrule
            \bottomrule
            \end{tabular} }}
    \end{center}
    \vspace{-0.1in}
\end{table}
\subsection{Ablation on Perturbation Type of Input}
\renewcommand{\arraystretch}{1.1}
\renewcommand{\tabcolsep}{4pt}
\begin{wraptable}[9]{r}{0.5\textwidth}
    \centering
    \vspace{-0.2in}
    \caption{\small Comparisons of perturbation type applied to the input. All models are adversarially-trained on CIFAR-10.}
    \label{tbl:ablation_input_perturb}
    \vspace{-0.1in}
    \resizebox{\linewidth}{!}{
    \begin{tabular}{lccccca}
    \toprule
     & \multicolumn{6}{c}{Adversarially-Trained} \\
     \cmidrule{2-7}
     & & PGD & FGSM & \multicolumn{2}{c}{HRS} & \cellcolor{white} \\
     \cmidrule(l{2pt}r{2pt}){3-3}\cmidrule(l{2pt}r{2pt}){4-4}\cmidrule(l{2pt}r{2pt}){5-6}
     & Clean & $\epsilon=1$ &  $\epsilon=8$ & PGD & FGSM & Avg. \\
    \midrule
    Gaussian Noise & 0.718  & 0.503 & 0.415& 0.590 & 0.603 & 0.566 \\
    Adversarial & 0.723 & 0.501 &  0.417 & 0.588 & 0.602 & 0.566  \\
            
    \midrule
    \bottomrule
    \end{tabular}}
\end{wraptable}
\label{sec:ablation_input_perturbation}
CRoZe evaluates the consistency between clean and perturbed inputs, regardless of the perturbation type, assuming that perturbed inputs retain the same semantic information as clean inputs. To empirically demonstrate the independence of our proxy from specific perturbation types applied to the input, we introduce random Gaussian noise in place of adversarial perturbations in Eq.~\ref{equation:attack_input} for obtaining our proxy values. As evidenced in Table~\ref{tbl:sup_input_perturbation_others}, CRoZe consistently exhibits similar Spearman's $\rho$ between the final performance of the standard-trained models and our proxy value in the NAS-Bench-201 search space on multiple benchmarks including CIFAR-10, CIFAR-100, and ImageNet16-120. Specifically, the gap of the average Spearman's correlation between the CRoZe using adversarial perturbations and the one with random Gaussian noise is merely 0.001 on CIFAR-100. Furthermore, when we measure the Spearmans' $\rho$ between the final performance of adversarially-trained models and our proxy, the average Spearman's $\rho$ values are the same between the CRoZe with random Gaussian noise and the one with adversarial perturbations (Table~\ref{tbl:ablation_input_perturb}). This emphasizes that CRoZe captures the characteristics of robust architectures through the consistency of clean and perturbed inputs, irrespective of the perturbation types employed.


\subsection{Ablation of the Weight Initialization Type}
\begin{table}[t]
\vspace{-0.1in}
    \begin{center}
    \caption{\small Comparison of Spearman's $\rho$ between the final accuracies and the proxy values on the NAS-Bench-201 search space with various weight initialization methods. All models are trained with both standard and adversarial training on CIFAR-10.}
    \label{tbl:sup_weight_init}
    \resizebox{0.9\linewidth}{!}{
        \small {
        \begin{tabular}{lccccaccccccca}
            \toprule
            \centering
             & \multicolumn{5}{c}{Standard-trained} & \multicolumn{8}{c}{Adversarially-trained}\\ 
             \cmidrule{2-6} \cmidrule{7-14}
             Weight Initialization Type & Clean & FGSM & PGD & CC. & Avg. & FGSM & PGD & CW & DeepFool & SPSA & LGV  & AutoAttack & Avg. \\ 
            \midrule
            \midrule
            Random & 0.823 & 0.826 & 0.188 & 0.436 & 0.568 & 0.441 & 0.532 & 0.220  & 0.454 & 0.240 &0.449& 0.458 & 0.399 \\
            Kaiming~\citep{he2015delving} & 0.812 & 0.818 & 0.189 & 0.430 & 0.562 & 0.428 & 0.512 & 0.217  & 0.443 & 0.227 & 0.426 & 0.436 & 0.384 \\
            Xavier~\citep{glorot2010understanding} & 0.816 & 0.822 & 0.190 & 0.433 & 0.565 & 0.428 & 0.513 & 0.217  & 0.442 & 0.227 & 0.425& 0.436 & 0.384  \\
            \midrule
            \bottomrule
            \end{tabular} }}
    \end{center}
    \vspace{-0.1in}
\end{table}
As our proxy assesses the robustness of the neural architecture within a single gradient step, we further investigate the sensitivity of our proxy to various weight initialization types. To validate the compatibility of our proxy with the diverse weight initialization type, we perform experiments employing Random initialization, Kaiming initialization~\citep{he2015delving}, and Xavier initialization~\citep{glorot2010understanding} on the NAS-Bench-201 search space on CIFAR-10. We measure Spearman's $\rho$ between the final accuracies and the proxy values, where the final performances are obtained by evaluating both standard-trained models and adversarially-trained models. We used 15,625 standard-trained models and 300 adversarially-trained models to ensure precise robustness assessment. Standard trained models are validated on clean images, adversarially-perturbed images (i.e., FGSM with an attack size of $8.0/255.0$ and PGD with an attack size of $1.0/255.0$), and 15 different types of common corrupted images. In contrast, advesrsarially-trained models are subjected to evaluation against 7 different types of strong adversarial attacks, including FGS, PGD, CW, DeepFool, SPSA, LGV, and AutoAttack. The results reported in the main paper are based on Random initialization. 

As demonstrated in Table~\ref{tbl:sup_weight_init}, our proxy maintains a consistently higher correlation compared to the baselines, irrespective of the weight initialization methods employed. Specifically, our proxy achieves an average correlation of 0.568 and 0.399 for standard training and adversarial training scenarios, respectively, whereas the best-performing baseline method only achieves 0.529 and 0.352 against various perturbations with the same random weight initialization. Since our approach considers the consistency of the parameters and gradients between the clean and perturbed images, our superior performance can be achieved regardless of the weight initialization method.

\subsection{Assessment of CRoZe Predictiveness on Adversarially-Trained Models}
\begin{figure}[h]
    \centering
    \caption{\small Spearman's $\rho$ in our proxy and each consistency component between the neural architectures with single-step trained states and fully-trained states against clean and perturbed images. All models are trained only on adversarially-perturbed images.}
    \label{fig:sup-proxy_consistency_at}
    \begin{subfigure}[b]{0.24\linewidth}
     \centering
         \includegraphics[width=\linewidth]{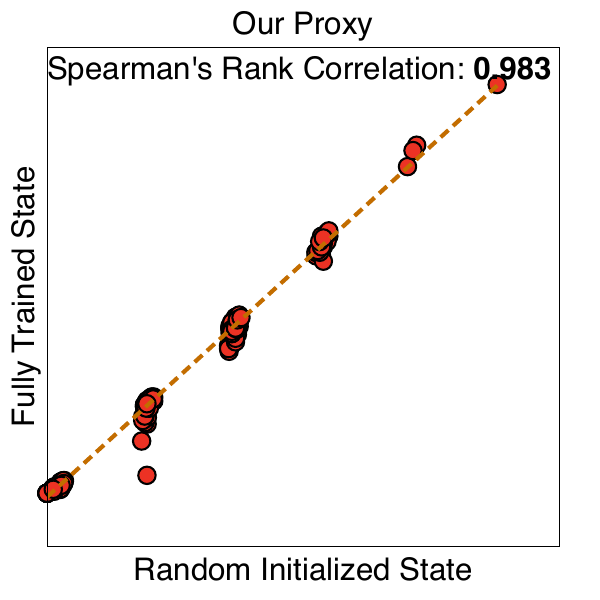}
         \vspace{-0.15in}
         \vspace{-0.1in}
         \label{fig:proxysim_at}
     \end{subfigure}
     \begin{subfigure}[b]{0.24\linewidth}
         \centering
         \includegraphics[width=\linewidth]{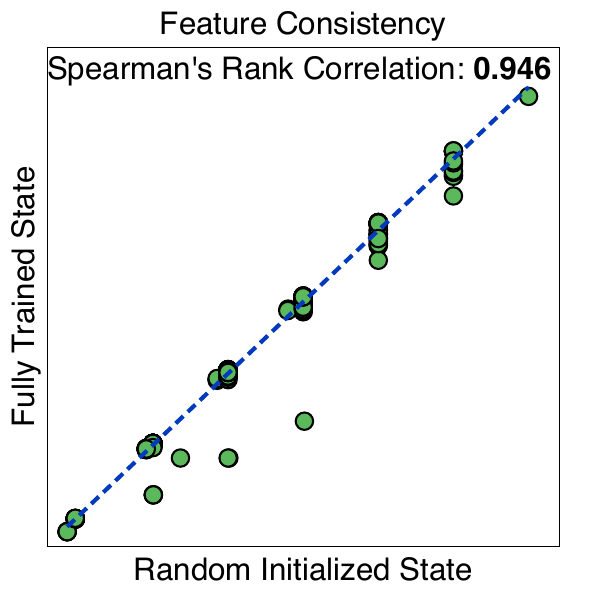}
         \vspace{-0.15in}
         \vspace{-0.1in}
         \label{fig:featsim_at}
     \end{subfigure}
     \begin{subfigure}[b]{0.24\linewidth}
         \centering
         \includegraphics[width=\linewidth]{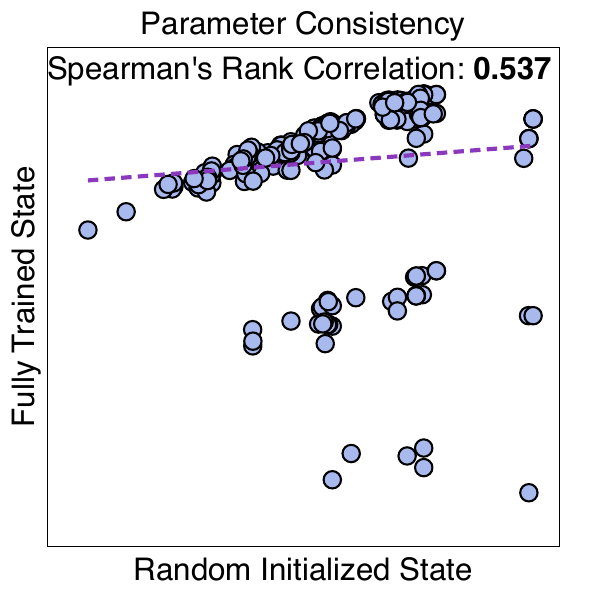}
         \vspace{-0.15in}
         \vspace{-0.1in}
         \label{fig:parmsim_at}
     \end{subfigure}
     \begin{subfigure}[b]{0.24\linewidth}
         \centering
         \includegraphics[width=\linewidth]{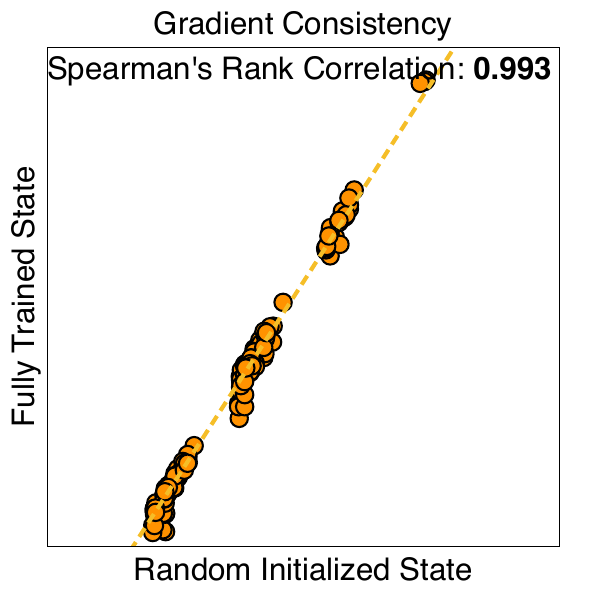}
         \vspace{-0.15in}
         \vspace{-0.1in}
         \label{fig:gradsim_at}
     \end{subfigure}
\end{figure}

As we verify the predictiveness of CRoZe with 300 standard-trained models in the NAS-Bench-201 search space in Section~\ref{sec:predictiveness}, we further validate the predictiveness of our proxy against adversarially-trained models. To achieve adversarial robustness, we adversarially train 300 randomly sampled architectures from the NAS-Bench-201 search space on the entire CIFAR-10 dataset, following~\citep{wong2020fast}. We then compute Spearman's $\rho$ between the values obtained from the fully-adversarially-trained states and single-step trained states associated with our proxy and each consistency component. 

Surprisingly, our proxy demonstrates a strong correlation of 0.983 between the proxy value obtained from a fully-trained state and a single-step trained state (Figure~\ref{fig:sup-proxy_consistency_at}). This indicates that our proxy accurately predicts the robustness of architectures even when they are adversarially-trained. Furthermore, each individual component of the proxy also exhibits a high correlation, suggesting that a strong predictiveness of our proxy against diverse perturbations does not rely on a dominant component, but rather on the overall precise evaluation provided by the combination of components.

\subsection{Versatility of CRoZe}
\begin{table}[ht]
\vspace{-0.1in}
    \begin{center}
    \caption{\small Validation of orthogonality of CRoZe. Comparison of Spearman's $\rho$ between the actual accuracies and the proxy values on CIFAR-10, CIFAR-100, and ImageNet16-120 in the NAS-Bench-201 search space. CC. stands for the average of 15 different types of common corruption. All models are standard-trained.}
    \label{tbl:sub_ensemble}
    \resizebox{0.9\linewidth}{!}{
        \small {
        \begin{tabular}{lccccccccccc}
            \toprule
            \centering
              & \multicolumn{4}{c}{CIFAR-10} & \multicolumn{4}{c}{CIFAR-100} & \multicolumn{3}{c}{ImageNet16-120} \\
            \cmidrule(r){2-5}\cmidrule(r){6-9}\cmidrule(r){10-12}
             Proxy Type& Clean & FGSM & PGD & CC. & Clean & FGSM & PGD & CC. & Clean & FGSM & PGD \\
            \midrule
            \midrule
            CRoZe & 0.823&0.826&0.188&0.823&0.784&0.786&0.343&0.784&0.765&0.596&0.707 \\
            Ensemble & 0.803&0.681&0.543&0.771&0.793&0.739&0.444&0.798&0.749&0.638&0.296 \\
            CRoZe+Ensemble&0.894&0.872&0.633&0.894&0.894&0.851&0.415&0.878&0.810&0.688&0.259 \\
            \midrule
            \bottomrule
            \end{tabular} }}
    \end{center}
    \vspace{-0.25in}
\end{table}
To demonstrate the versatility of our robust zero-shot proxy, CRoZe, and its compatibility with existing clean zero-shot NAS approaches~\citep{abdelfattah2021zero}, we conduct additional experiments where we ensemble our proxy with other methods. Following~\citet{abdelfattah2021zero}, we re-implemented ensemble-based NAS, which involves aggregating predictions from multiple zero-shot NAS methods to calculate the proxy score for a given neural architecture. For our ensemble, we select \{Snip, NASWOT\} as the baseline. We then compare 1) CRoZe, 2) Ensemble: \{Snip, NASWOT\}, and 3) CRoZe+Ensemble: \{CRoZe, Snip, NASWOT\} in the NAS-Bench-201 search space on CIFAR-10, CIFAR-100, and ImageNet16-120. Robust accuracies are obtained by evaluating the standard-trained models against FGSM attack with an attack size of $8.0/255.0$ and PGD with an attack size of $1.0/255.0$. 

The results presented in Table~\ref{tbl:sub_ensemble} indicate that CRoZe significantly enhances the predictiveness for both clean and robust accuracies, encompassing adversarial attacks and common corruption across all tasks. Notably, CRoZe+Ensemble showcases substantial improvements, achieving increases of 11.33\% in clean accuracy and 28.05\%, 16.57\%, and 15.95\% in robustness concerning FGSM attack, PGD attack, and common corruption, respectively on CIFAR-10. These results underscore the effectiveness of our proxy when combined with other proxies, enabling a more precise robust neural architecture search.

\subsection{End-to-End Performance on ImageNet16-120}

\begin{table}[t]
\centering
\caption{\small Comparisons of the final performance and required computational resources of the searched neural architectures in the DARTS search space on ImageNet16-120. CC. stands for the average of 15 different types of common corruption.}
\label{tbl:sup-e2e_imagenet} 
\vspace{0.05in}
\resizebox{0.7\textwidth}{!}{
		\begin{tabular}{lccccccca}
	   	\toprule
	   	
	   	\multicolumn{1}{c}{\multirow{2}{*}{NAS Method}} & 
	   	Training-free                                         &
	   	Params                                      & 
	   	\multirow{2}{*}{\# GPU}                      & 
            \multirow{2}{*}{Batch Size}                 &
	   	\multicolumn{4}{c}{Standard-Trained}        \\
            \cmidrule{6-9}
	   	                                          & 
	      NAS                                   & 
	   	(M)                                         & 
	   	                                          &
                                                        &
            Clean                                       &
            CC.                                         &
            FGSM                                        &
            HRS                                         \\
	   	\midrule
	   	\midrule
	   	PC-DARTS~\citep{xu2019pc}                   &
	   	                                        &
	   	5.30                                        &
	   	8                                           &
            1024                                        &
	   	50.58                                       &
	   	14.36	   		   	                        &
            0.15                                        &
            0.30                                        \\
	   	DrNAS~\citep{chen2020drnas}                 &
	   	                                          &
	   	5.70                                        &
	   	8                                           &
            512                                         &
	      49.63                                       &
	      13.42 	   		   	                        &
            0.21                                         &
            0.42                                        \\
            AdvRush~\citep{mok2021advrush}              &
	   	                                           &
	   	4.20                                        &
	   	1                                           &
            64                                          &  
	      38.72                                       &
	      10.39	   		   	                        &
            0.11                                        &
            0.22                                       \\
            GradNorm~\citep{abdelfattah2021zero}         &
	   	\checkmark                                  &
	   	5.90                                        &
	   	1                                           &
            8                                           &
            39.13                                       &
            10.75                                       &
	   	0.23                                         &
	   	0.46                                       \\
            SynFlow~\citep{abdelfattah2021zero}         &
	   	\checkmark                                  &
	   	6.13                                        &
	   	1                                           &
            8                                           &
            43.73                                       &
            12.10                                       &
	   	0.15                                        &
	   	0.30                                        \\
            CRoZe                                        &
	   	\checkmark                                  &
	   	5.87                                        &
	   	1                                           &
            8                                           &
	   	47.90                                       &
	   	13.35            		   	                &
            0.32                                        &
            0.64                                         \\
	   	\midrule
	   	\bottomrule
        \end{tabular}}
        \vspace{-0.1in}
\end{table}

We further validate the final performance of the searched neural architectures by CRoZe and compare the required computational resources with existing NAS frameworks including robust NAS (AdvRush~\citep{mok2021advrush}), clean one-shot NAS (PC-DARTS~\citep{xu2019pc}, DrNAS~\citep{chen2020drnas}) and clean zero-shot NAS (SynFlow and GradNorm~\citep{abdelfattah2021zero}), on ImageNet16-120 in the DARTS search space. Similar to Section~\ref{sec:e2e_darts}, we sample the same number (e.g., 5,000) of candidate architectures using the warmup+move strategy with an init pool of 3,000 and sample pool of 50 for both clean-zero shot NAS and CRoZe. 

The NAS Training-free methods such as GradNorm, SynFlow, and CRoZe only require a single GPU with a batch size of 8 to search for the architectures on the ImageNet16-120 dataset. In contrast, the existing clean one-shot NAS methods require 8 GPUs with much larger batch sizes. Moreover, NAS-Training-free methods consume less than 3000MB of memory, while both clean one-shot NAS and robust NAS need at least 3090 RTX GPU, which is available at 24000MB of memory. With its superior computational efficiency, CRoZe enables rapid neural architecture search and achieves the best HRS accuracy while maintaining comparable clean and common corruption accuracies. all at a much lower computational cost (Table~\ref{tbl:sup-e2e_imagenet}). These demonstrate the effectiveness of CRoZe for rapid and lightweight robust NAS across diverse tasks (i.e., CIFAR-10, CIFAR-100, and ImageNet16-120) and perturbations (i.e., adversarial perturbations and common corruptions).

\end{document}